\documentclass[lettersize,journal]{IEEEtran}
\usepackage{amsmath,amsfonts}
\usepackage{bbm}
\usepackage{array}
\usepackage[caption=false,font=normalsize,labelfont=sf,textfont=sf]{subfig}
\usepackage{textcomp}
\usepackage{stfloats}
\usepackage{url}
\usepackage{verbatim}
\usepackage{graphicx}
\usepackage{cite}

\usepackage{colortbl}
\usepackage{xcolor}
\usepackage{makecell}
\usepackage{enumitem}
\usepackage{multirow}
\usepackage{pifont}

\usepackage[colorlinks=true]{hyperref}
\usepackage[capitalize]{cleveref}

\usepackage[dvipsnames]{xcolor}

\newcommand{\bc}{\mathbf{c}}
\newcommand{\bd}{\mathbf{d}}

\newcommand{\bo}{\mathbf{o}}
\newcommand{\bp}{\mathbf{p}}

\newcommand{\br}{\mathbf{r}}
\newcommand{\bs}{\mathbf{s}}

\newcommand{\bx}{\mathbf{x}}

\newcommand{\bSigma}{\boldsymbol{\Sigma}}

\newcommand{\cG}{\mathcal{G}}

\newcommand{\cP}{\mathcal{P}}

\makeatletter
\DeclareRobustCommand\onedot{\futurelet\@let@token\@onedot}
\def\@onedot{\ifx\@let@token.\else.\null\fi\xspace}
\def\eg{e.g\onedot} 
\def\ie{i.e\onedot}

\makeatother

\definecolor{yellow}{rgb}{1, 1, 0.7}
\definecolor{orange}{rgb}{1, 0.85, 0.7}
\definecolor{tablered}{rgb}{1, 0.7, 0.7}
\definecolor{red}{rgb}{1, 0, 0}

\definecolor{wincolor}{rgb}{0.85, 0.0, 0.0}

\definecolor{darkyellow}{rgb}{0.8, 0.8, 0.5}
\definecolor{darkred}{rgb}{0.7, 0.3, 0.3}
\definecolor{darkgreen}{rgb}{0.3, 0.7, 0.3}
\definecolor{blue}{rgb}{0.251, 0.498, 0.824}
\definecolor{green}{rgb}{0, 1.0, 0}
\definecolor{pink}{rgb}{1, 0.4, 0.7}

\definecolor{realred}{rgb}{0.95, 0.1, 0.0}

\usepackage{algorithm}
\usepackage{algpseudocode}

\def\eg{\textit{e.g.}}
\def\ie{\textit{i.e.}}

\definecolor{yellow}{rgb}{1,1, 0.7}
\definecolor{lightyellow}{rgb}{1,1, 0.8}
\definecolor{orange}{rgb}{1, 0.85, 0.7}
\definecolor{tablered}{rgb}{1, 0.7, 0.7}

\definecolor{tabthird}{rgb}{1, 1, 0.7}
\definecolor{tabsecond}{rgb}{1, 0.85, 0.7}
\definecolor{tabfirst}{rgb}{1, 0.7, 0.7}

\newcommand{\revise}[1]{\textcolor[rgb]{0, 0, 0}{#1}}

\begin{document}



\title{{Efficient Scene Modeling via Structure-Aware and Region-Prioritized 3D Gaussians}}

\author{Guangchi Fang, Bing Wang\textsuperscript{$\ast$}
\thanks{
This work is jointly supported by Young Scientists Fund of the National Natural Science Foundation of China (42301520), Research Grants Council of Hong Kong (25206524).

Guangchi Fang and Bing Wang are with The Spatial Intelligence Group, The Hong Kong Polytechnic University, HKSAR.
Corresponding author: Bing Wang (bingwang@polyu.edu.hk).
}
}



\maketitle

\begin{abstract}


Reconstructing 3D scenes with high fidelity and efficiency remains a central pursuit in computer vision and graphics. 
Recent advances in 3D Gaussian Splatting (3DGS) enable photorealistic rendering with Gaussian primitives, yet the modeling process remains governed predominantly by photometric \revise{supervision.}
\revise{This reliance} often leads to irregular spatial \revise{distribution} and indiscriminate primitive adjustments that \revise{largely ignore} underlying geometric context. 
In this work, we rethink Gaussian modeling from a geometric standpoint and introduce Mini-Splatting2, \revise{an efficient scene modeling framework} that couples structure-aware distribution and region-prioritized optimization, \revise{driving} 3DGS into a geometry-regulated paradigm.
\revise{The structure-aware distribution} enforces spatial regularity through structured \revise{reorganization} and representation sparsity, ensuring balanced structural coverage for compact organization.
\revise{The region-prioritized optimization} improves \revise{training discrimination} through geometric saliency and computational selectivity, fostering \revise{appropriate} structural emergence for fast convergence.
These mechanisms \revise{alleviate} the long-standing tension among representation compactness, convergence acceleration, and rendering fidelity.
Extensive experiments demonstrate that Mini-Splatting2 achieves up to 4$\times$ fewer Gaussians and 3$\times$ faster optimization while maintaining state-of-the-art visual quality, \revise{paving the way towards} structured and efficient 3D Gaussian modeling.


\end{abstract}


{
\section{Introduction}
\label{sec:intro}

\IEEEPARstart{T}{he} growing demand for compact, scalable, and high-fidelity 3D scene modeling is driving progress across domains such as autonomous systems \cite{zhou2024drivinggaussian, song2025adgaussian}, digital twins \cite{Huang2DGS2024, Yu2024GOF}, mixed reality \cite{matsuki2024gaussian, yan2024gs}, and immersive content creation \cite{tang2023dreamgaussian, LaRa}.
These scenarios require not only visually accurate scene representations, but also modeling frameworks that are memory-conscious, computationally lightweight, and amenable to fast training and deployment.
While neural rendering approaches \cite{mildenhall2020, barron2021mipnerf, barron2023zip} have achieved impressive results under image supervision, their reliance on high-capacity dense representations incurs substantial computational overhead, limiting their suitability for real-time and large-scale use.
This highlights the need for scene modeling paradigms that balance accuracy with efficiency, enabling expressive yet lightweight representations that scale across tasks and environments.


Among recent advances, 3D Gaussian Splatting (3DGS) \cite{kerbl20233d} has emerged as a promising alternative for novel view synthesis, representing scenes as collections of spatially anchored and view-dependent Gaussians.
This formulation combines the geometric interpretability of explicitly parameterized primitives with the representational flexibility of optimizable attributes, enabling parallelized rasterization and fast convergence.
Despite these advantages, the scene modeling process still exhibits inefficiencies in both \revise{spatial distribution} and training dynamics, particularly under multiview photometric supervision without explicit geometric context.
These inefficiencies hinder modeling scalability and quality, motivating a closer investigation into their fundamental causes.


\begin{figure}[t]
	\centering
	\includegraphics[width=1\linewidth]{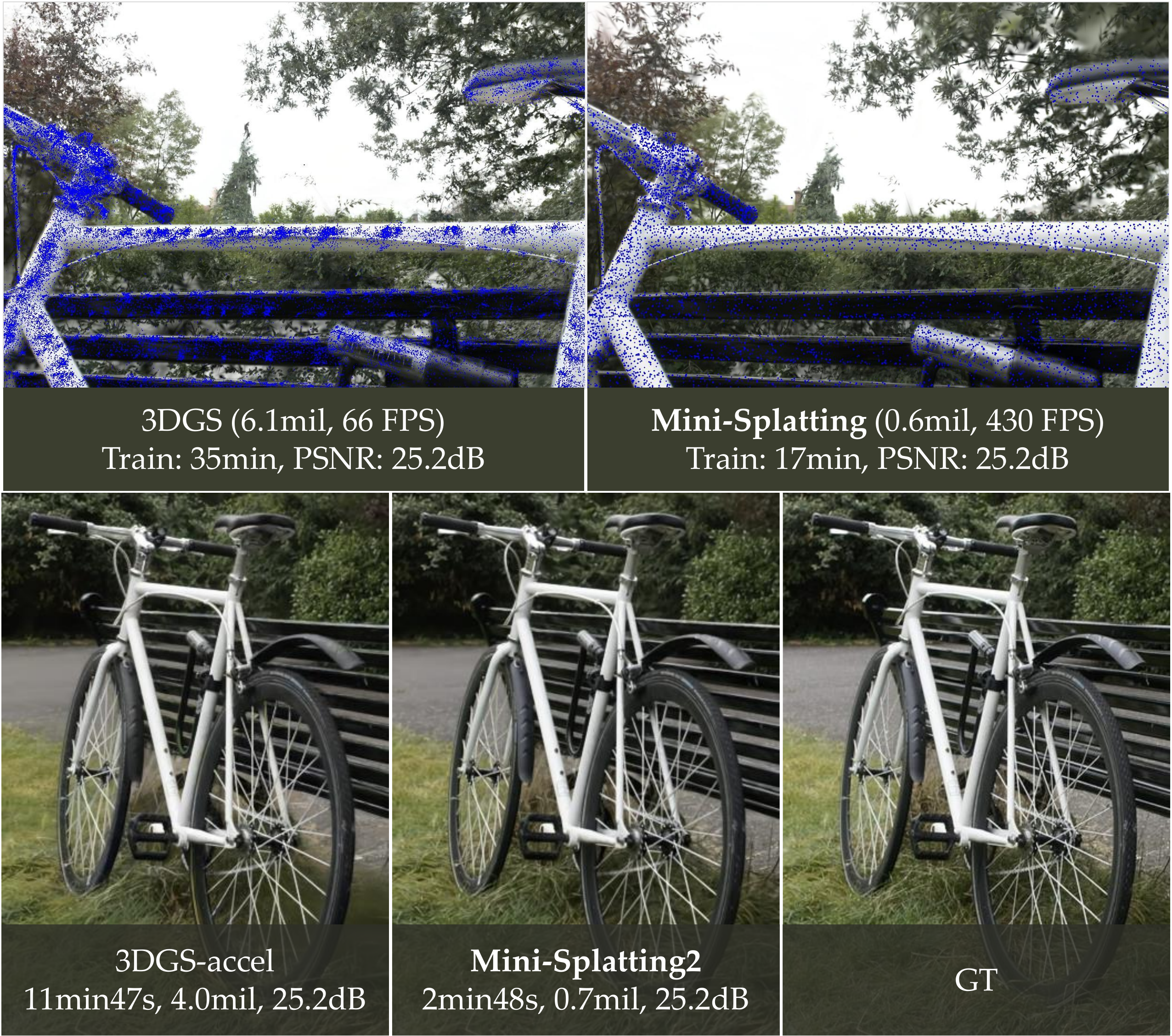}
    \caption{
Gaussians in 3DGS exhibit a highly irregular spatial distribution, which inherently limits its representational capability.
To address this, Mini-Splatting introduces a structure-aware distribution scheme that mitigates such irregularity, substantially reducing the number of Gaussians while improving rendering speed, training efficiency, and visual quality. 
Building on this foundation, Mini-Splatting2 incorporates a region-prioritized optimization scheme to further accelerate training. 
In a comparative evaluation on the \textit{bicycle} scene, Mini-Splatting2 demonstrates substantial improvements over the latest official accelerated 3DGS implementation, referred to as 3DGS-accel~\cite{kerbl20233d, taming3dgs}, requiring 5.7$\times$ fewer Gaussians and achieving a 4.2$\times$ speedup in optimization, while maintaining state-of-the-art visual quality.
    }
	\label{fig:teaser}
\end{figure}


A key source of inefficiency lies in the absence of geometric principles governing how Gaussians are spatially managed. 
In standard 3DGS pipelines, the spatial distribution of Gaussians is adjusted through an adaptive density control strategy, which measures the positional gradient magnitude of Gaussians to trigger threshold-based split, clone, or prune operations. 
This locally gradient-driven process can adaptively refine high-contrast regions, often geometric edges or high-frequency textures, while reducing redundancy in smooth areas, such as plain walls and flat floors. 
However, \revise{without considering Gaussians as the essential underlying structure}, this strategy over-concentrates on those with high positional gradients and under-represents structurally important yet low-contrast regions, as illustrated in Fig.~\ref{fig:teaser}. 
Such irregularity results in imbalanced scene coverage, degraded reconstruction fidelity, increased rasterization cost, and inflated memory consumption.

A parallel inefficiency arises from the absence of geometric criteria guiding how Gaussians are dynamically optimized. 
In most existing 3DGS pipelines, the optimization process is driven solely by photometric consistency across views, typically by per-pixel loss metrics (\eg, PSNR) and structural similarity measures (\eg, SSIM). 
To maintain stability under such weak supervision, these pipelines adopt conservative progressive growth schedules, incrementally increasing the number of Gaussians over time, and apply uniform updates to all Gaussians, regardless of their visibility or contribution to the rendered views. 
However, without a geometric relevance assessment across Gaussians, they cannot effectively trigger early activation or guide targeted attribute updates of structurally informative regions. 
Such indiscrimination disregards regionally geometric priority, wastes computation on occluded or uninformative Gaussians, delays coherent scene structure emergence, and prolongs the optimization process.



Inspired by these sources of inefficiency, we propose Mini-Splatting2, a rearchitected 3DGS framework that improves modeling performance through \textbf{structure-aware distribution} and \textbf{region-prioritized optimization}.
\revise{Our approach is built on two core assumptions.
First, imposing spatial regularity on Gaussians in regions with pronounced image gradients ensures balanced coverage, thereby supporting compact organization.
Second, applying training discrimination to geometrically informative areas accelerates structural emergence, thus promoting fast convergence.}
Our method introduces two mechanisms to jointly regulate the \revise{spatial distribution} and training dynamics of Gaussian primitives, thereby improving both modeling efficiency and reconstruction fidelity.


To mitigate the irregularity in \revise{spatial distribution}, we propose a structure-aware distribution scheme that explicitly regulates the spatial arrangement of Gaussians. 
This scheme comprises two synergistic modules that jointly improve representation efficiency. 
First, \textbf{adaptive Gaussian organization} improves coverage balance \revise{through blur split and depth reinitialization.
Blur split} identifies low-contrast and blurry regions that lack sufficient coverage and adaptively increases their local density.
\revise{Depth reinitialization} consolidates multi-view depth estimates to reposition Gaussians in under-reconstructed regions.
Second, \textbf{redundant Gaussian simplification} reduces memory overhead \revise{via intersection preserving and primitive sampling.
Intersection preserving} models ray-mesh intersections to retain structurally critical Gaussians.
\revise{Primitive sampling} introduces stochastic sampling to maintain a sparse yet informative set of primitives.
In combination, these modules arrange Gaussians into a spatially coherent distribution, lowering both memory usage and rasterization cost without sacrificing reconstruction fidelity.


To alleviate the indiscrimination in training dynamics, we propose a region-prioritized optimization scheme that steers the optimization process toward geometrically salient regions of the scene. This scheme consists of two complementary components that collectively accelerate convergence.
First, \textbf{aggressive model growth} hastens early structure emergence \revise{through critical Gaussian identification and aggressive Gaussian clone.
Critical Gaussian identification} pinpoints primitives with high reconstruction significance.
\revise{Aggressive Gaussian clone} selectively duplicates them to rapidly expand scene coverage in the initial optimization stage.
Second, \textbf{occluded Gaussian culling} reduces update overhead \revise{via visibility identification and view-dependent culling.
Visibility identification} quantifies per-view primitive visibility derived from blended rendering weights, serving as a proxy for geometric relevance.
\revise{View-dependent culling} discards occluded or low-visibility Gaussians during rasterization, avoiding unnecessary updates.
Working in tandem, these components reconfigure standard discriminative optimization into a geometry-prioritized process, achieving faster convergence while preserving rendering quality and maintaining modeling compactness.

Our main contributions are as follows:
\begin{itemize}
\setlength{\itemsep}{0pt}
\setlength{\parsep}{0pt}
\setlength{\parskip}{0pt}

\item  
We provide a \revise{comprehensive} analysis of 3D Gaussians from a geometric perspective and identify two key inefficiencies, irregular \revise{spatial distribution} and indiscriminate training dynamics, that limit modeling compactness and convergence. 
These insights motivate Mini-Splatting2, a unified framework for efficient scene modeling that improves both memory and computational efficiency.

\item 
We propose a structure-aware distribution scheme to enhance representation compactness by enforcing spatial regularity among Gaussians.
It integrates adaptive Gaussian organization to address coverage imbalance and redundant Gaussian simplification to eliminate superfluous Gaussians. 
This design yields compact and expressive representations, thereby reducing memory consumption.

\item 
We present a region-prioritized optimization scheme to accelerate convergence by emphasizing geometrically discriminative regions. 
It combines aggressive model growth for rapid structural emergence and occluded Gaussian culling to suppress less informative updates. 
This scheme delivers a targeted and streamlined optimization process, thus reducing computational overhead.

\item 
Extensive experiments on various public benchmarks demonstrate that Mini-Splatting2 achieves a \textbf{4$\times$} reduction in Gaussian numbers and delivers a \textbf{3$\times$} acceleration in optimization time, while preserving rendering quality comparable to state-of-the-art 3DGS implementations. 
These results \revise{position it as a strong and practical baseline} for efficient real-world scene modeling.



\end{itemize}

A preliminary version of this work was published at ECCV 2024 as Mini-Splatting \cite{fang2024mini}. This extended version introduces substantial methodological and empirical advancements beyond the original. 
First, we provide a \revise{comprehensive} analysis of the geometry modeling process in 3DGS, offering new insights into the structural capacity of Gaussians and demonstrating their potential for dense point cloud reconstruction. 
Second, we propose a region-prioritized optimization scheme that incorporates two effective acceleration mechanisms, aggressive model growth and occluded Gaussian culling, which significantly enhance optimization efficiency by emphasizing geometrically informative regions. 
Third, we integrate these advances into a unified and rearchitected framework, Mini-Splatting2, which delivers a favorable trade-off among representation compactness, optimization speed, and rendering fidelity. This extension reflects a major expansion in technical contribution and experimental validation. \href{https://github.com/fatPeter/mini-splatting2}{Code Available}.

}

\section{Related Work}
\label{sec:related_work}

\subsection{Gaussian Splatting}

3D Gaussian Splatting (3DGS) \cite{kerbl20233d} was originally developed for novel view synthesis, with a strong emphasis on real-world immersive rendering. Early methods in this domain \cite{mildenhall2020} employed multi-layer perceptrons to implicitly model scenes as neural radiance fields. 
More recent approaches have moved towards explicit representations, including voxel grids \cite{liu2020neural, sun2022direct}, hash grids \cite{muller2022instant}, and point clouds \cite{kopanas2021point, xu2022point}, which enhance representation capacity and consequently improve both training efficiency and rendering quality. Expanding on classical 3D representations, elliptical Gaussians \cite{zwicker2002ewa} offer robust capabilities for modeling scene geometry and appearance while inherently supporting parallel rasterization \cite{kerbl20233d}. This approach eliminates the need for ray marching \revise{required by} previous volume rendering techniques~\cite{mildenhall2020}, thereby accelerating both optimization and rendering.

Owing to the strong scalability of Gaussian representations, 3DGS has demonstrated promising potential across a wide range of applications. Studies from diverse fields, including 3D reconstruction, computer graphics, and robotics, have applied this work to downstream tasks such as surface reconstruction \cite{Huang2DGS2024, Yu2024GOF}, content generation \cite{tang2023dreamgaussian, LaRa}, and localization and mapping \cite{matsuki2024gaussian, yan2024gs}. While several studies have sought to improve 3DGS from the perspectives of graphics \cite{franke2024trips, mai2024ever}, image processing \cite{yu2023mip}, and system performance \cite{durvasula2023distwar, taming3dgs}, relatively few have explored the fundamental characteristics of 3D Gaussians as a scene representation.

Despite notable progress in novel view synthesis, existing approaches still depend on dense representations and superfluous optimization, which limit their practicality in real-world applications such as robotics, visualization, and simulation. 
As both academic and industrial demands evolve, there is an increasing need for lightweight Gaussian representations and fast Gaussian optimization to effectively meet performance constraints in practical deployments.

\subsection{Lightweight Gaussian Representation}

With recent advancements in neural radiance fields, several studies have focused on simplifying neural representations, primarily \revise{centering on} feature grids and introducing voxel pruning \cite{deng2023compressing, li2023compressing} or voxel masking \cite{xie2023hollownerf, rho2023masked}. 
Similar to these developments, Gaussian pruning has been explored in recent 3DGS compression-oriented research \cite{niedermayr2024compressed, lee2024compact, fan2024lightgaussian, navaneet2024compgs, ali2024trimming, zhang2025gaussianspa}. 
For instance, C3DGS \cite{niedermayr2024compressed} proposed pruning Gaussians based on image gradients, while Compact-3DGS \cite{lee2024compact} and LightGaussian \cite{fan2024lightgaussian} applied pruning based on Gaussian opacity and scale. However, these approaches overlook the issue of irregular spatial distribution of Gaussians, which can lead to suboptimal simplification results following pruning.

Another line of research focuses on enhancing data structures to achieve more lightweight scene representations. Scaffold-GS \cite{lu2024scaffold} introduced structured 3D Gaussians with anchor points and multi-layer perceptrons, and subsequent studies \cite{chen2024hac, wang2024contextgs} have further investigated compact representations through advanced compression approaches. Other studies~\cite{kerbl2024hierarchical, ren2024octree, liu2024citygaussian} have integrated Octree structures and Level-of-Detail techniques with Gaussian representations, enhancing their applicability to large-scale scene modeling. Nonetheless, these approaches generally neglect the inherent correlation between Gaussians and points, and thus fail to fully exploit geometric guidance from the underlying scene structure.

\begin{figure*}[tb]
	\centering
	\includegraphics[width=0.8\linewidth]{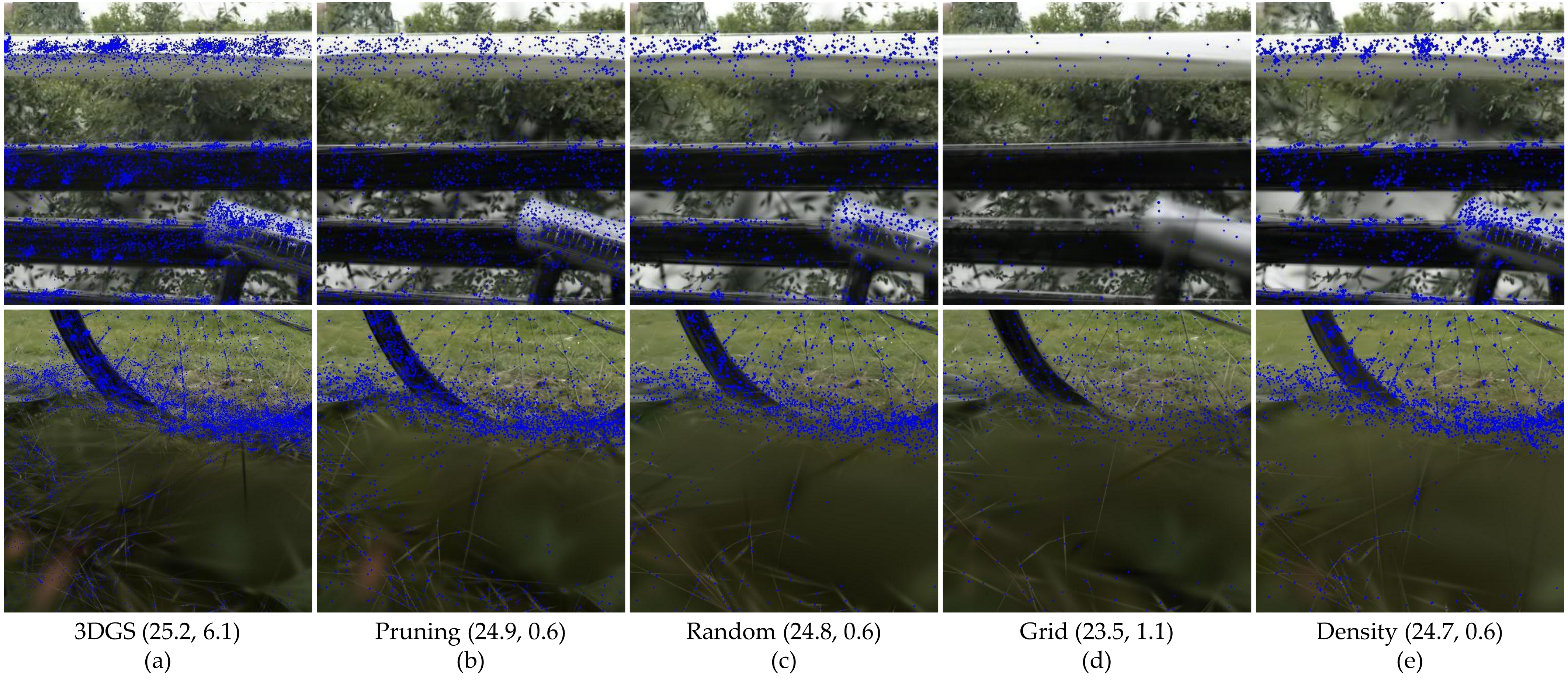}
    \caption{Analysis of Gaussian Representation. (a) Projected Gaussian centers in the vanilla 3DGS, along with the corresponding rendering quality (PSNR in dB) and total number of Gaussians (in millions). Notable issues of `overlapping' and `under-reconstruction' are clearly observed. (b)-(e) Visualization of Gaussian centers after applying different simplification strategies, including pruning, random sampling, grid sampling, and density-preserved sampling, respectively.
    }
	\label{fig:Visual_analysis}
\end{figure*}

\subsection{Fast Gaussian Optimization}

Gaussian densification plays a crucial role in recovering scene geometry, directly influencing Gaussian optimization. The vanilla 3DGS approach \cite{kerbl20233d} initializes \revise{Gaussians} with sparse points from Structure-from-Motion (SfM) \cite{schoenberger2016sfm} and employs adaptive density control that first selects Gaussians based on image-space gradients and then clones or splits them according to their scale. Building on this approach, recent studies \cite{kim2024color, zhang2024pixelgs, ye2024absgs, bulo2024revising, taming3dgs} have focused on Gaussian selection strategies informed by image-space information and intrinsic Gaussian attributes. Other approaches incorporate multi-view constraints \cite{cheng2024gaussianpro, li2025mvg, du2024mvgs} or leverage advanced optimization techniques~\cite{kheradmand20243d} to enhance densification. A well-designed densification strategy can improve optimization convergence, thereby accelerating the overall optimization process. However, most existing methods primarily aim to enhance rendering quality, often leading to an increased optimization budget.

A parallel line of research \cite{fan2024instantsplat, charatan23pixelsplat, chen2024mvsplat, liu2025mvsgaussian, szymanowicz2024flash3d} explores learning-based approaches to directly predict scene geometry in the form of depth maps, dense point clouds, or Gaussian attributes. These methods have demonstrated significant potential for 3D reconstruction, particularly in sparse-view scenarios and structured environments. 
However, in the general and classical multi-view setting, still the dominant configuration for practical 3D scene reconstruction, learning-based approaches remain in an early stage, facing challenges such as maintaining geometric consistency and robust multi-view alignment.

A promising direction in the classical multi-view setting is the enhancement of both forward and backward rasterization pipelines~\cite{durvasula2023distwar, ye2024gsplat, taming3dgs}. For example, DISTWAR~\cite{durvasula2023distwar} accelerates atomic operations through warp-level reduction. In addition, the codebase of Taming-3DGS \cite{taming3dgs}, referred to as 3DGS-accel, incorporates several optimization components, including per-splat backward rendering, fused-SSIM computation, spherical harmonic separation, and tile-based culling \cite{radl2024stopthepop}, resulting in a reported 2.7$\times$ speedup in training time. Collectively, these works provide strong baselines for 3DGS and have advanced its development and applications. However, further improvements can be realized by incorporating geometric criteria to guide Gaussian optimization more effectively.

\vspace{0.5cm}
{Compared with concurrent works, which achieve lightweight representation and accelerated optimization primarily through advanced densification and optimization strategies, our method explicitly adopts a geometric perspective rather than relying solely on multi-view photometric supervision. In particular, our approach is distinguished in two ways: 1) it treats the spatial positions of Gaussians as an integral component of the underlying scene structure, thereby mitigating irregularity in \revise{spatial distribution}; and 2) it evaluates geometric relevance across Gaussians to establish regionally prioritized geometry, reducing indiscrimination during training dynamics.}

\section{Geometric Analysis within Gaussian Splatting}
\label{sec:preliminaries}
\newcommand{\RRR}{\mathbb{R}}

\subsection{Preliminaries of Gaussian Splatting}\label{sec:Preliminaries of Gaussian Splatting}

\noindent\textbf{Gaussian Representation.} 3DGS \cite{kerbl20233d} represents scene geometry as a collection of 3D Gaussians, denoted as $\{\cG_i | i=1, \cdots, N\}$. Each Gaussian $\cG_i$ is characterized by its opacity $\alpha_i\in[0,1]$, center $\bp_i \in \RRR^{3 \times 1}$ and covariance matrix in world space $\bSigma_i \in \RRR^{3\times3}$ \cite{zwicker2002ewa}, formulated as $\cG_i(\bx) = e^{-\frac{1}{2} (\bx-\bp_i)^T \bSigma_i^{-1}(\bx-\bp_i)}$. To render an image from Gaussians, 3DGS first sorts them in an approximate depth order based on the distance from their centers $\bp_i$ to the image plane, and further applies alpha-blending \cite{mildenhall2020} as follows:
\begin{equation}\label{eq:render} 
    \bc(\bx) = \sum_{i=1}^N w_i \bc_i, \quad \text{where } w_i=T_i \alpha_i  \cG^{2D}_i(\bx).
\end{equation}
In this formulation, $\bc_i$ denotes the view-dependent color modeled by spherical harmonic (SH) coefficients. The accumulated transmittance $T_i$ is defined as $T_i=\prod_{j=1}^{i-1} (1 - \alpha_j \cG^{2D}_j(\bx))$, where $\cG^{2D}_i$ represents the 2D Gaussian projection of $\cG_i$, obtained via a local affine transformation \cite{zwicker2002ewa}.

\noindent\textbf{Adaptive Density Control.} 3DGS employs an adaptive density control strategy to recover scene geometry. Starting from sparse SfM points as the initial Gaussians, 3DGS dynamically adjusts the density of the representation by adding and removing Gaussians during optimization. Specifically, Gaussians exhibiting significant image-space positional gradients are either cloned or split, depending on their scales, to improve local detail. Conversely, Gaussians with low opacity or excessive screen-space size are pruned to avoid redundancy and preserve efficiency. Ablation studies in~\cite{kerbl20233d} demonstrate that this densify-and-prune strategy adaptively grows the number of Gaussians for higher-quality scene modeling.

\subsection{Irregularity within Gaussian Representation}\label{sec:Challenges}

In this section, we first analyze the irregular spatial distribution of Gaussians resulting from the adaptive density control strategy. Then, we demonstrate that straightforward Gaussian simplification, which disregards their role as underlying geometry, is insufficient to address this irregularity.
\revise{Finally, we analyze the necessity of enforcing spatial regularity to achieve compact organization within Gaussian representation.}

\noindent\textbf{Analysis of Gaussian Representation.} Adaptive density control evaluates the positional gradient magnitude of Gaussians to trigger threshold-based densification.
Specifically, during the optimization process, each Gaussian accumulates positional gradients according to the photometric optimization objective. 
During densification, Gaussians whose accumulated gradients exceed a predefined threshold are selected to undergo \revise{split or clone operations} for increased density. 
Nevertheless, this threshold-based selection may introduce spatial irregularity when Gaussians are either over- or under-selected due to suboptimal threshold settings, as shown in Fig.~\ref{fig:Visual_analysis} (a).

In textureless regions, characterized by homogeneous appearance and negligible image gradient magnitudes, 3DGS suppresses the density of Gaussians, which is beneficial for scene modeling.
During optimization, Gaussians accumulate positional gradients from the photometric objective, which become extremely small in regions with negligible image gradients.
During densification, adaptive density control skips such Gaussians, and distributes them sparsely within these regions, thereby reducing redundancy.
Consequently, 3DGS avoids densifying Gaussians in color-smooth areas, such as plain walls or floors, and instead employs a small number of large Gaussians to capture both structure and appearance.
This strategy reduces the number of Gaussians in such regions without compromising rendering quality.


In contrast, textured regions, characterized by pronounced image gradient magnitudes, exhibit spatial irregularity within Gaussian representation, reflecting both redundancy and insufficiency.
For regions with high image gradients, the corresponding Gaussians accumulate large positional gradients, prompting adaptive density control to densely select Gaussians, which leads to over-densification.
Conversely, in regions with relatively low image gradients, this strategy tends to ignore Gaussians, behaving similarly to textureless regions, which can result in under-representation.
This strategy induces spatial irregularity in textured regions, as illustrated in Fig.~\ref{fig:Visual_analysis} (a). 
The first row highlights regions of `overlapping', where dense clustering of Gaussians occurs in areas with significant color variance. When alpha blending is applied for rendering, these overlapping regions incur redundant computation. 
The second row depicts `under-reconstruction', where regions with relatively low image gradients are insufficiently covered, leading to blurry artifacts. 
These observations reveal the irregular distribution of Gaussians and underscore the inherent redundancy and insufficiency in the scene representation.


\noindent\textbf{Gaussian Pruning and Sampling.} A straightforward strategy for constructing a lightweight Gaussian representation is pruning and sampling. Following the importance estimation approach in \cite{li2023compressing}, we compute an importance score for each Gaussian $\cG_i$ by accumulating its blending weights across all intersecting rays as $I_i = \sum_{j=1}^K w_{ij}$, where $K$ denotes the total number of contributing rays. Gaussians with importance scores below a predefined threshold are pruned, and the resulting representation is further optimized until convergence. This approach considers only blending weights during image rendering as photometric guidance while ignoring the spatial positions of Gaussians as indicators of the underlying geometry. Therefore, it cannot fully address spatial irregularity. As shown in Fig.~\ref{fig:Visual_analysis} (b), this method reduces the number of Gaussians from 6.1 million to 0.6 million. However, it also leads to a degradation in rendering quality, with PSNR dropping from 25.2~dB to 24.9~dB, and fails to resolve the issues of `overlapping' and `under-reconstruction'.


To further investigate the impact of 3D simplification, we integrated pruning with several standard point cloud sampling strategies, including random sampling, grid sampling, and density-preserved sampling, as illustrated in Figs.~\ref{fig:Visual_analysis} (c), (d), and (e), respectively. 
Although these methods successfully simplify Gaussians to varying extents, they still struggle to address the irregular spatial distribution.




\noindent\revise{
\textbf{Spatial Regularity.} 
In textureless regions, where image gradients are negligible, spatial regularity is not necessarily required, as a small number of Gaussians can adequately model such regions. 
In contrast, in textured regions, which exhibit pronounced image gradients, enforcing spatial regularity is desirable to mitigate the irregular distribution of Gaussians, thereby ensuring a compact and expressive representation.
}

\begin{figure}[tb]
	\centering
	\includegraphics[width=1\linewidth]{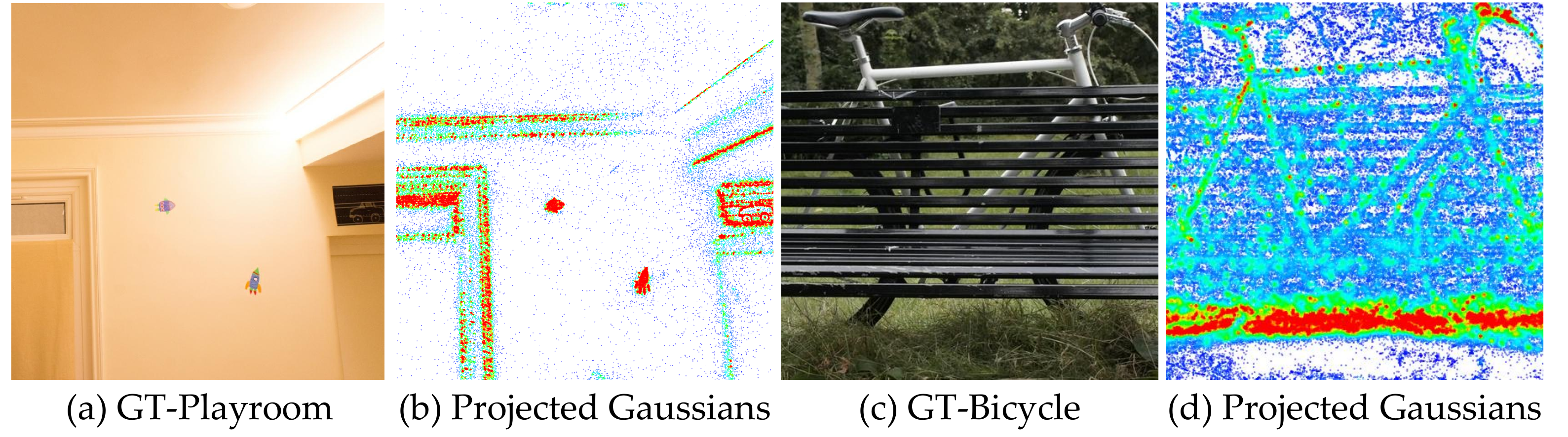}
    \caption{
    \revise{
    Spatial irregularity. (a, c) Ground-truth images of the \textit{playroom} and \textit{bicycle} scans. (b, d) Corresponding projected Gaussian centers, colored by density from low (blue) to high (red).
    }
    }
	\label{fig:regularity}
\end{figure}

\revise{
In textureless regions, the image gradients are negligible, and when minimizing the photometric optimization objective between the rendered and input views, they yield low positional gradients for the associated Gaussians.
These positional gradients rarely exceed the predefined densification threshold, leaving the small set of initial Gaussians.
Since these regions are largely color-smooth, these sparse Gaussians are sufficient for maintaining rendering quality while remaining memory-compact.
Therefore, spatial regularity, which promotes a balanced distribution, is not strictly necessary in this case.
Figure \ref{fig:regularity} (b) illustrates that 3DGS avoids densifying Gaussians in textureless regions, where a small number of large Gaussians adequately model these areas.
This result indicates that spatial regularity is not necessary in textureless regions.
}

\revise{
In textured regions, enforcing spatial regularity is desirable. 
Textured regions with high image gradients yield large positional gradients through the optimization objective.
With these large positional gradients, Gaussians are repeatedly selected for densification, leading to clustering of Gaussians.
Conversely, textured regions with moderately low image gradients produce relatively small positional gradients.
Similar to textureless regions, these low gradients do not trigger densification, leading to insufficient coverage with sparse Gaussians.
Such clustering introduces redundant computation, while insufficient coverage leads to blurry artifacts, collectively highlighting the inherent inefficiency of the representation.
Therefore, spatial regularity is necessary to mitigate both excessive clustering and insufficient coverage.
As further illustrated in Fig. \ref{fig:regularity} (d), the foreground bicycle exhibits pronounced Gaussian clustering, whereas adjacent regions suffer from inadequate coverage, being represented by only a small number of Gaussians.
These observations indicate that spatial regularity is desirable for textured regions.
}

\revise{
Motivated by this analysis, our approach seeks to promote spatial regularity in textured regions by explicitly regulating the spatial distribution of Gaussians.
This encourages a more balanced arrangement, reducing redundant Gaussians and mitigating blurry artifacts, thereby ensuring a compact and expressive representation.
A more detailed discussion of spatial distribution is provided in Sec. \ref{Sec:Discussion}.
}

\begin{figure}[tb]
	\centering
	\includegraphics[width=1\linewidth]{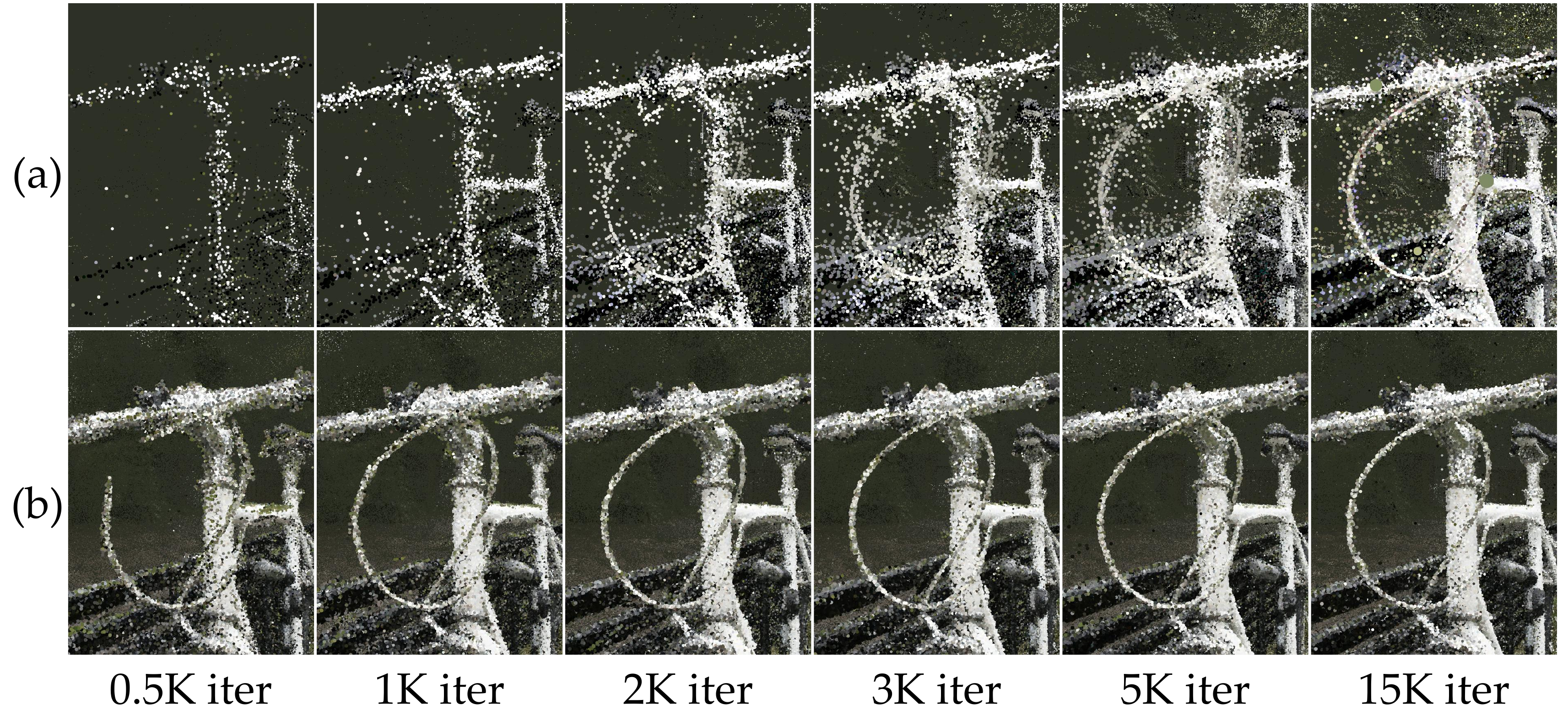}
    \caption{Analysis of Gaussian Optimization. (a) Gaussian centers. (b) Corresponding depth points across 0.5K to 15K iterations. While the Gaussian centers exhibit some visual artifacts, a dense and informative point cloud can still be effectively extracted from the Gaussian representation.
    }
	\label{fig:geom}
\end{figure}

\begin{figure*}[tb]
	\centering
	\includegraphics[width=0.8\linewidth]{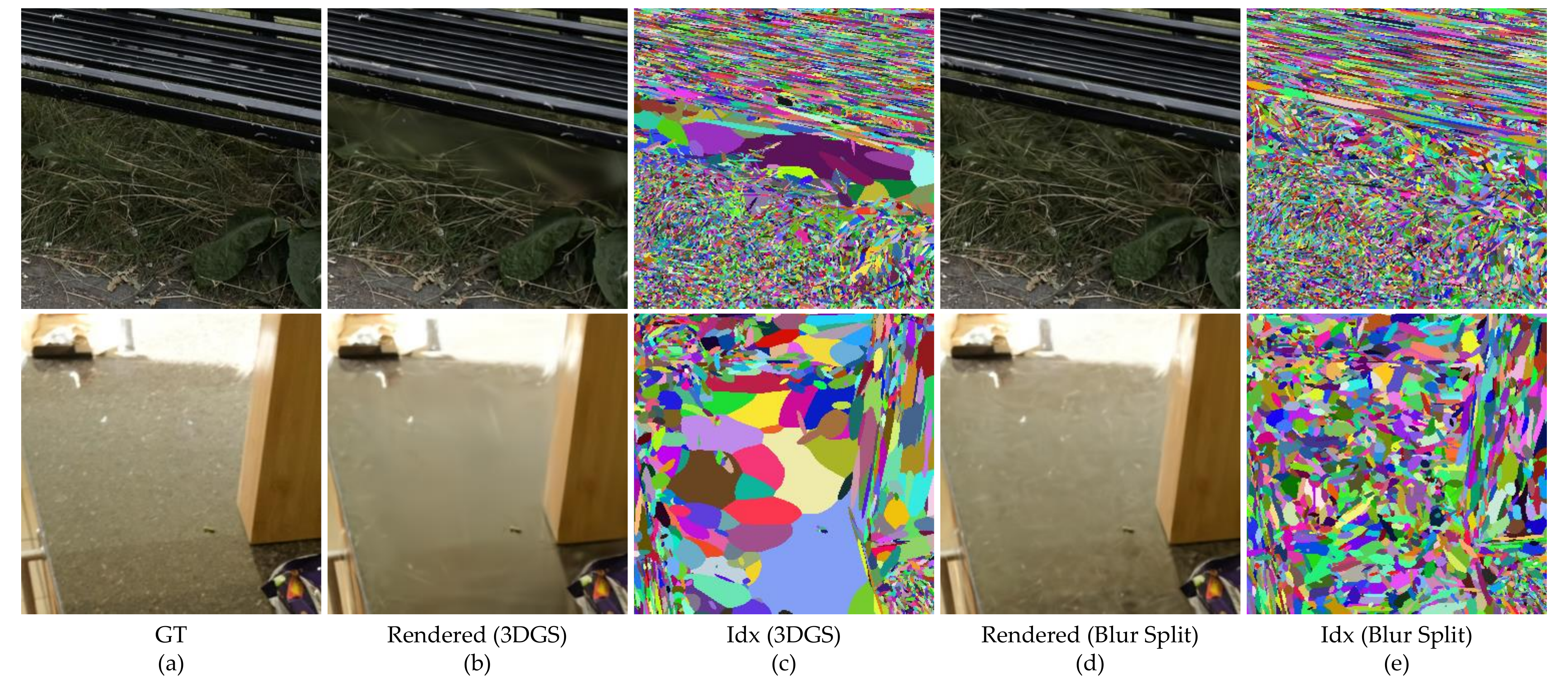}
    \caption{Visual analysis of the blur split component. (a) Ground truth image. (b) Image rendered using the original 3DGS \cite{kerbl20233d}. (c) Gaussian indices corresponding to the maximum contribution for each pixel. (d) Image rendered after applying blur split. (e) Gaussian indices after applying blur split.
    } 
	\label{fig:blur_split}
\end{figure*}

\subsection{Indiscrimination within Gaussian Optimization}\label{sec:Reconstruction}

In this section, we analyze the optimization progress of Gaussians optimized solely through photometric consistency. We demonstrate that this process inherently encodes latent geometric cues, which in turn provide regionally geometric priors to guide discriminative optimization.

\noindent{\textbf{Analysis of Gaussian Optimization.} Existing 3DGS pipelines primarily rely on photometric loss to drive the optimization, which induces indiscrimination in optimization. To ensure stability under weak supervision metrics such as PSNR and SSIM, these pipelines gradually increase the number of Gaussians while applying uniform updates across all instances. However, such a strategy fails to promote early activation, delaying the emergence of coarse scene structures, and does not prioritize structurally informative regions, thereby expending unnecessary computation on occluded Gaussians.}

Figure~\ref{fig:geom} (a) illustrates the typical optimization progress in 3DGS. Starting from an initial sparse set of points (0.5K iterations), the number of Gaussians progressively increases through iterative optimization and densification, resulting in a coarse approximation of scene geometry with substantial floaters between 0.5K and 5K iterations. With further optimization, Gaussians converge toward object surfaces (up to 15K iterations), though a subset persists as floaters to account for camera- or view-dependent effects. \revise{Typically}, the presence of incomplete structures with significant floaters, particularly at an early stage of optimization (\ie, 1K to 3K iterations), indicates suboptimal reconstruction quality.

\noindent\textbf{Point Cloud Reconstruction via Gaussians.} The latent geometric cues within Gaussians can be explicitly extracted to mitigate training indiscrimination. In particular, point cloud reconstruction can be derived by leveraging geometry-related Gaussian attributes within a depth estimation framework. Building upon the image rendering formulation in \cref{eq:render}, we define the reconstructed depth map $d$ as:
\begin{equation}\label{eq:render_depth} 
    d(\bx) = d^{mid}_{i_{max}}(\bx), \quad \text{where } i_{max}=\arg\max\limits_{i} w_i.
\end{equation}
Here, $d^{mid}$ denotes the depth derived from ray-Gaussian ellipsoid intersections, and $w_i$ is the blending weight as defined in \cref{eq:render}. The resulting depth maps can be reprojected into world coordinates to yield reconstructed point clouds.

A key observation is that dense point clouds can be effectively reconstructed even in the early stages of optimization using this approach. As shown in Fig. \ref{fig:geom} (b), Gaussian representations with imperfect spatial positions (\eg, at 1K to 3K iterations) are still capable of producing reasonable geometric reconstructions, particularly for foreground objects. This observation reveals a latent geometric prior embedded in the training dynamics of Gaussians and highlights the potential to accelerate convergence by steering the optimization process toward geometrically salient regions of the scene.

\section{{Structure-Aware Distribution Scheme for Gaussian Representation}}
\label{sec:method1}

To regulate their spatial distribution, we treat Gaussians as an essential component of the underlying structure and introduce a structure-aware distribution scheme that comprises two components, adaptive Gaussian organization, which improves scene coverage balance, and redundant Gaussian simplification, which reduces memory consumption. These components collectively mitigate irregularity in \revise{spatial distribution}.


\subsection{{Adaptive Gaussian Organization}}\label{sec:Densification}

In contrast to the adaptive density control strategy, which adjusts the number of Gaussians, our adaptive Gaussian organization introduces two key components, blur split and depth initialization, that adaptively organize the spatial locations of Gaussians, thereby improving scene coverage balance.

\noindent\textbf{Blur Split.} Our blur split component is designed to increase the local density of blurry regions with insufficient coverage, as shown in Fig. \ref{fig:blur_split} (b), thereby mitigating the `under-reconstruction' issue discussed in Sec. \ref{sec:Challenges}. The gradient-based adaptive density control strategy proposed in \cite{kerbl20233d} may fail in areas with smooth color transitions, as illustrated in Fig. \ref{fig:blur_split} (a). Consequently, the corresponding oversized Gaussians $\cG_i$ tend to be retained during optimization. An intuitive implementation adopted in 3DGS is to prune Gaussians with excessively large screen-space scales based on their projected 2D distribution $\cG^{2D}_i$. However, this pruning method does not effectively resolve the issue of `under-reconstruction'.

As illustrated in Fig. \ref{fig:blur_split} (c), we render indices of Gaussians with the maximum contribution to a pixel as $i_{max}=\arg\max\limits_{i} w_i$ (\ie, the Gaussian with the maximum weight in alpha-blending) onto the image. The rendered indices \( i_{max}(\bx) \) for each pixel \( \bx \) are mapped to random colors for visualization. To \revise{distinguish from} the rendered indices corresponding to Gaussians with the maximum contribution $i_{max}(\bx)$, we denote the projected index of the $i$th Gaussian $\cG_i$ as $i(\bx)$. A key observation is that, compared to the projected index $i(\bx)$ (\ie, the projected area of Gaussian $\cG_i$), the blurry artifacts are more directly associated with the rendered index $i_{max}(\bx)$. This suggests that Gaussians with large maximum contribution areas contribute to the problem of `under-reconstruction'. 

\begin{figure*}[tb]
	\centering
	\includegraphics[width=0.8\linewidth]{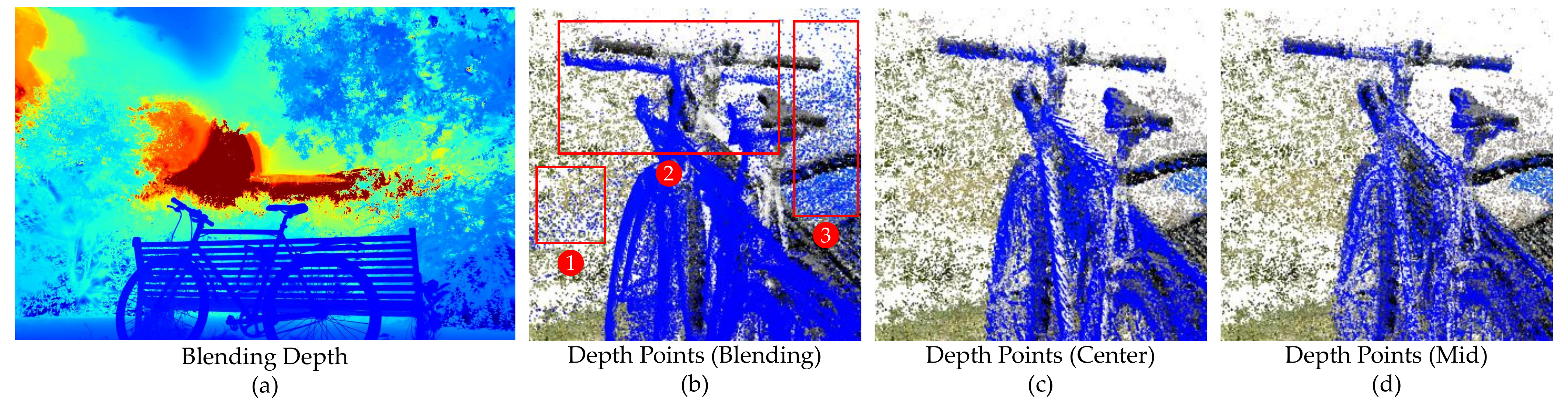}
    \caption{Visual analysis of depth maps and reconstructed depth points from 3DGS \cite{kerbl20233d}. (a) Depth map rendered via alpha blending using the original 3DGS model. (b) Depth points reconstructed from the alpha-blended depth map, where red boxes indicate significant geometric artifacts and inconsistencies, especially around object boundaries and fine structures. (c) Depth points reconstructed from our predicted depth map, where the depth values are anchored at the centers of the corresponding Gaussians, resulting in more accurate and smoother geometry. (d) Depth points reconstructed from our predicted depth map using midpoints along the Gaussian ray intersections, offering further geometric refinement and ensuring continuity.}
	\label{fig:depth_points}
\end{figure*}



Building upon this observation, we identify Gaussians with large blurry regions for each input image, denoted as $\cG^{blur}$, using the following criterion:
\begin{equation}\label{eq:blur_split} 
    \cG^{blur}=\{\cG_i | S_i > \mathcal{T}_{blur} \land i \in [1, N]\}, \quad \mathcal{T}_{blur}=\theta_{blur} \cdot H \cdot W.
\end{equation}
Here, $S_i=\sum_{\bx=(1,1)}^{(H, W)} \delta(i(\bx)=i_{max}(\bx))$ represents the maximum contribution area of Gaussian $\cG_i$, where $\delta$ is the indicator function, and $(H, W)$ denotes the image resolution. The value of $S_i$ can be efficiently computed during the forward pass of rasterization. The hyperparameter $\theta_{blur}$ is empirically set to $2 \times 10^{-4}$ to control the threshold for identifying blurry regions. Gaussians satisfying the criterion for $\cG^{blur}$ are then split according to the strategy in \cite{kerbl20233d} during the optimization process. The effectiveness of this deblurring method is demonstrated in Fig. \ref{fig:blur_split} (d) and (e), which show how our blur split component mitigates the `under-reconstruction' issue, thereby enhancing scene coverage and image clarity.

\noindent\textbf{Depth Reinitialization.} Our depth reinitialization component leverages dense depth information to reposition Gaussians, thereby mitigating both the `overlapping' and `under-reconstruction' phenomena in Sec. \ref{sec:Challenges}. The use of depth to enhance neural rendering is a well-established and effective strategy, as demonstrated in several prior NeRF-based algorithms~\cite{deng2022depth, xu2022point}. However, in 3DGS-based approaches, the primary challenge is not merely utilizing the reconstructed depth but rather generating accurate and practical depth data.

Similar to NeRF \cite{mildenhall2020}, 3DGS can render a blended depth map by replacing the Gaussian color $\bc_i$ with the depth of its center $d_i$, formulated as $d^{blend} = \sum_{i=1}^N w_i \cdot d_i$. This formulation has been widely adopted in subsequent works, and the resulting depth map is shown in Fig. \ref{fig:depth_points} (a). However, we observe that this seemingly reasonable 2D depth map can be misleading, often producing a depth point cloud with significant artifacts. In Fig.~\ref{fig:depth_points} (b), we reproject the blended depth into world space and visualize the Gaussian centers with their corresponding colors. Three representative failure cases, depth collapse, object misalignment, and blending boundary, are highlighted with red boxes for clarity. These issues primarily arise from the use of alpha blending across multiple overlapping Gaussians, which can lead to inconsistent depth estimations.

Here, we present our formulation for rendering depth. Given an elliptical Gaussian defined by its scale $\bs=(s_x, s_y, s_z)$, we model it as a 3D ellipsoid: $g(x, y, z)=\frac{x^2}{s_x^2} + \frac{y^2}{s_y^2} + \frac{z^2}{s_z^2} = 1$. For an input ray parameterized as $\br(t)=\bo+t\bd$, where $\bo$ and $\bd$ represent the ray origin and direction transformed into the coordinate system of the ellipsoid, we compute the midpoint between the two intersection points of the ray and the ellipsoid. The depth at this midpoint is defined as the Gaussian depth $d^{mid}_i$. An alternative approach would be to determine the optimal $t$ corresponding to the maximum density along the ray. We adopt the midpoint formulation, as it yields numerically identical results to the alternative method, while additionally providing a useful discriminant (\ie, a criterion for determining whether the ray intersects the ellipsoid).

\begin{figure}[tb]
	\centering
	\includegraphics[width=1\linewidth]{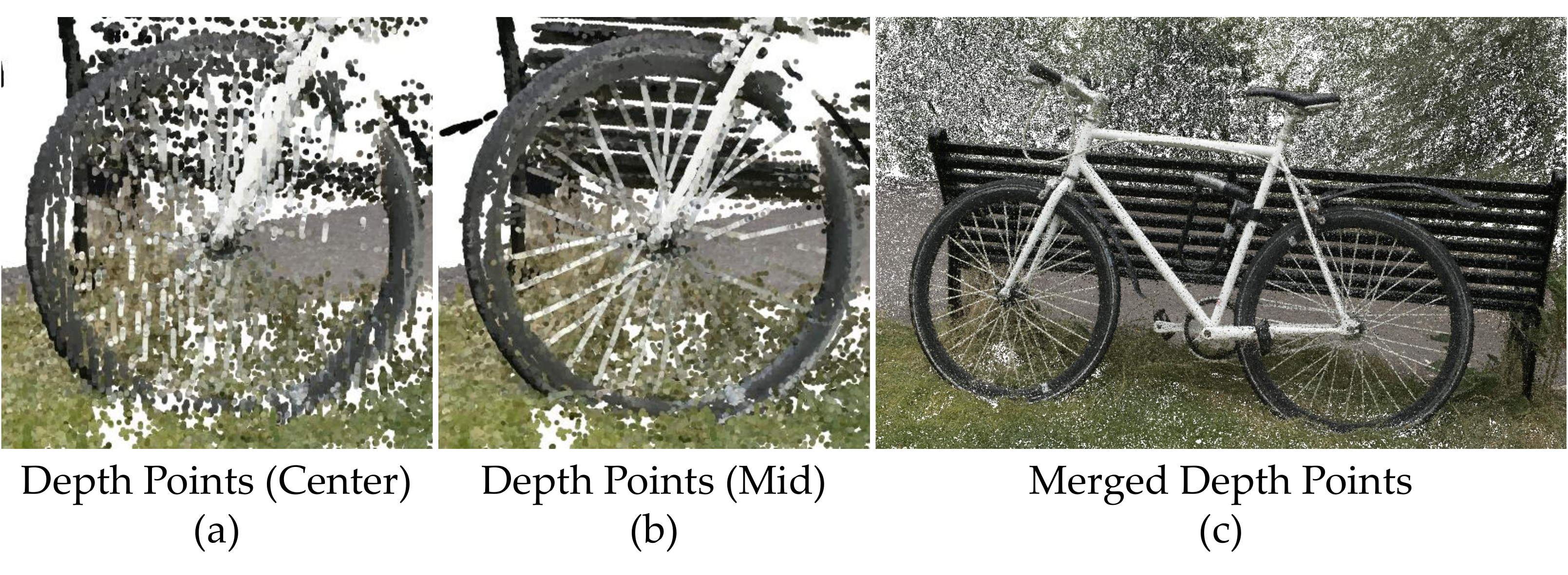}
    \caption{Depth points. (a) Depth points from our depth map using Gaussian centers. (b) Depth points from our depth map using midpoints. (c) Merged depth points from our depth map using midpoints. The Depth Points (Mid) exhibit improved point cloud reconstruction results.
    }
	\label{fig:merged_depth}
\end{figure}

In scenarios involving multiple Gaussians, we select the one with the maximum contribution to each pixel in order to mitigate artifacts caused by alpha blending. The final depth $d^{mid}$ is defined as $d^{mid}=d^{mid}_{i_{max}}$, where $i_{max}=\arg\max\limits_{i} w_i$. In practice, we observe that setting the depth based on the Gaussian center, as $d^{center}=d_{i_{max}}$, yields comparable rendering quality to $d^{mid}$ after further optimization. However, since $d^{mid}$ yields superior reconstruction results, as demonstrated in Figs.~\ref{fig:merged_depth} (a) and (b), we adopt it in our Mini-Splatting pipeline to support potential downstream applications.

After generating the depth map, we reproject all depth points into world space. For each image, a subset of points is randomly selected as initialization seeds, to which the corresponding ground-truth colors are assigned. During optimization, we repeatedly reinitialize our Gaussian representation with these points to explicitly organize the spatial distribution of Gaussians. The resulting merged point cloud is shown in Fig. \ref{fig:merged_depth} (c). These depth points facilitate more balanced scene coverage within our model, thereby mitigating the spatial irregularity inherent in the original representation.

\subsection{{Redundant Gaussian Simplification}}

Our redundant Gaussian simplification strategy consists of two key components, intersection preserving and primitive sampling, which jointly eliminate superfluous Gaussians, thereby reducing memory consumption while preserving structural fidelity and rendering quality.

\noindent\textbf{Intersection Preserving.} Inspired by ray-mesh intersection methods, our intersection preserving removes Gaussians that do not directly intersect with the rendering ray. Unlike hard-thresholding approaches, we avoid binarizing smooth opacity or blending weights into discrete ${0, 1}$ values, as doing so may degrade visual fidelity. Instead, our approach retains structurally critical Gaussians, thereby maintaining a balance between structural precision and rendering quality.

In depth reinitialization, we represent 3D Gaussians as ellipsoids and determine their depths based on midpoints along the viewing rays. We render the depth of the Gaussian with the maximum contribution, representing it as the depth of the scene. These midpoints, corresponding to the intersections of rays with the scene representation, serve as the foundation for the set of intersected Gaussians, denoted as $\cG^{int}$. 
Specifically, for each image, we define the set of intersected Gaussians $\cG^{int}$ as those that are the primary contributors to at least one pixel:
\begin{equation}
    \cG^{int}=\{\cG_{i} | i \in \mathcal{I}_{max} \land i \in [1, N]\},
\end{equation}
where $\mathcal{I}_{max}$ denotes the set of all indices within the rendered indices $i_{max}(\bx)$, and $N$ is the number of Gaussians.

\begin{figure}[tb]
	\centering
	\includegraphics[width=0.8\linewidth]{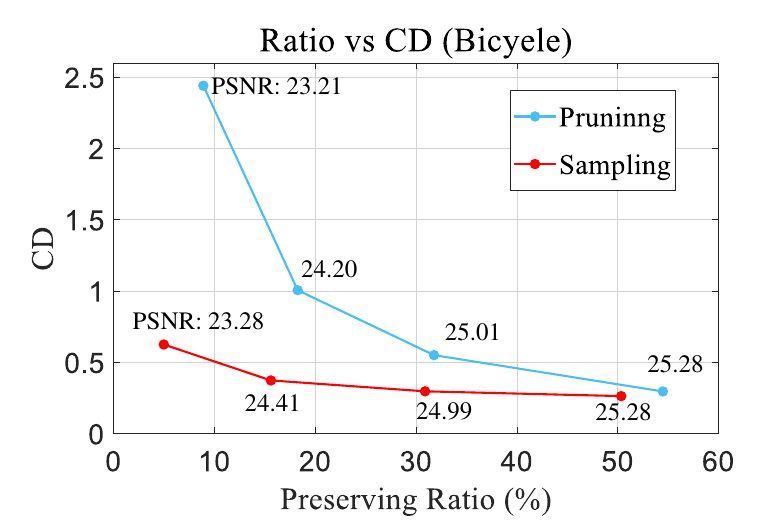}
    \caption{The relation between the preserving ratio and geometry. Our sampling strategy demonstrates improved geometric fidelity and rendering quality, particularly at low Gaussian preservation ratios.
    }
	\label{fig:sampling}
\end{figure}

\noindent\textbf{{Primitive Sampling.}} In this section, we analyze the limitations of direct Gaussian pruning and propose our primitive sampling. This component incorporates stochastic sampling informed by scene geometry, thereby maintaining rendering quality while preserving a sparse set of Gaussians.

The concept of pruning in neural representations was initially introduced in grid-based neural radiance fields \cite{yu2021plenoxels, li2023compressing} to eliminate empty or low-information voxels. This idea can be naturally extended to Gaussian representations by assigning each Gaussian a predefined importance score $I_i$ and pruning those with low values. While this approach is effective for removing a small subset of Gaussians, applying it with a low preservation ratio often results in a noticeable degradation of rendering quality. This limitation arises because the importance metric reflects significance only prior to simplification. Naive pruning may disrupt local geometric continuity, thereby impairing final rendering quality after optimization. In particular, neighboring Gaussians within a local region tend to share similar importance scores, which can lead to their simultaneous removal or retention, risking the loss of essential geometric structures. To assess this effect, we perform pruning on the set of intersected Gaussians and evaluate geometric fidelity using the Chamfer distance between the centers of the pruned and non-simplified Gaussian sets, as shown in Fig.~\ref{fig:sampling}. The substantial increase in Chamfer distance observed at low preservation ratios highlights that relying solely on Gaussian importance results in suboptimal simplification.

Compared to deterministic pruning, a stochastic sampling strategy offers improved preservation of overall geometric structures. Motivated by this observation, we incorporate Gaussian importance $I_i$ into the sampling probability of each Gaussian, defined as $\cP_i=\frac{I_i}{\sum_{i=1}^N {I_i}}$, which enables weighted sampling in proportion to the relative importance of each Gaussian. As shown in Fig.~\ref{fig:sampling}, primitive sampling more effectively retains local geometry, thereby enhancing the final rendering quality. In our experiments, we find that importance computed from blending weights consistently yields superior performance in both pruning and sampling. Furthermore, combining blending weight with additional attributes, such as the projected scale of Gaussians or the rendered index, leads to varying levels of effectiveness across indoor and outdoor scenes, highlighting the potential for the design of adaptive importance to different scans.

\section{{Region-Prioritized Optimization Scheme for Gaussian Optimization}}
\label{sec:method2}

To emphasize geometric priority during optimization, we focus on salient regions and introduce a region-prioritized optimization scheme comprising two key components, aggressive model growth, which hastens early structure emergence, and occluded Gaussian culling, which reduces unnecessary update overhead. Together, these components alleviate the indiscrimination in training dynamics.

\subsection{Aggressive Model Growth}\label{sec:Aggressive_Gaussian_Densification}

We first propose aggressive model growth as an alternative to the conventional progressive Gaussian \revise{growth} schedule in order to accelerate early structure emergence. 
This schedule can be achieved through Gaussian densification in an aggressive manner (\ie, aggressive Gaussian densification), which incorporates two key components, critical Gaussian identification and aggressive Gaussian clone. 
Together, these components enable the model to dramatically and selectively clone Gaussians with high reconstruction significance during the early stages of optimization.

\begin{figure}[tb]
	\centering
	\includegraphics[width=1\linewidth]{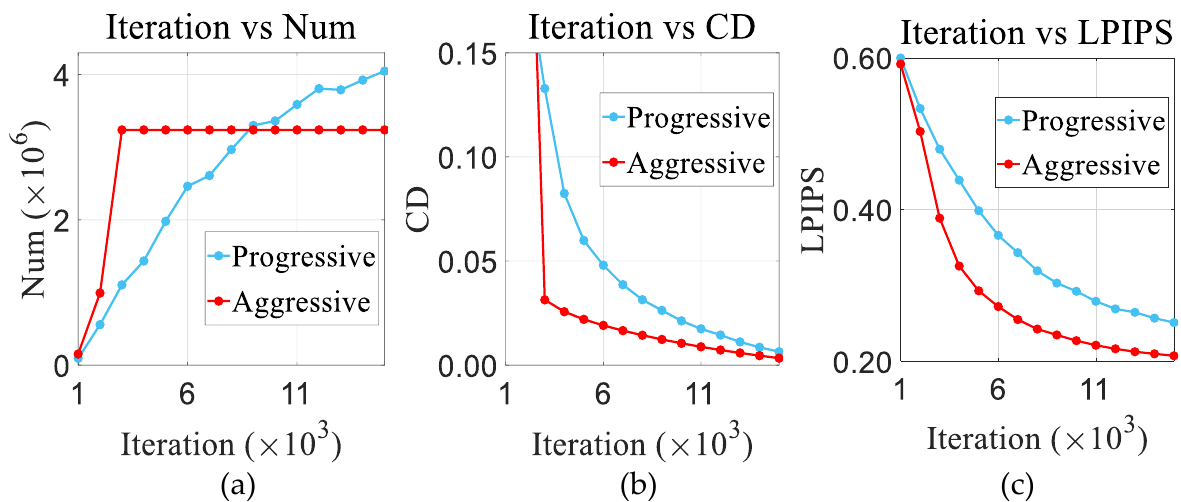}
    \caption{The relationship between the optimization progress of \textit{bicycle} and (a) the number of Gaussians, (b) the geometric similarity compared with resulting Gaussians, and (c) the rendering quality, as observed under the progressive growth and our aggressive growth patterns.}
	\label{fig:cd_psnr}
\end{figure}

\noindent\textbf{Model Growth from Progressive to Aggressive.} A typical Gaussian optimization process can be divided into two main phases: the densification phase (\ie, optimization coupled with Gaussian densification before the 15K iterations) and the refinement phase (\ie, pure optimization after the 15K iterations). 
Most existing approaches employ a progressive scheduler to grow the number of Gaussians following the principle of \textsl{{Progressive Model Growth}} adopted in 3DGS~\cite{kerbl20233d}, which incrementally increases the number of Gaussians in a restrained and iterative manner over an extended densification period, as illustrated in Fig.~\ref{fig:cd_psnr}.

In contrast to existing progressive patterns, we propose an aggressive schedule, motivated by the analysis in Sec.~\ref{sec:Reconstruction}. \textsl{{Aggressive Model Growth}} aims to rapidly densify 3D Gaussians within a constrained number of optimization iterations, thereby significantly shortening the densification period and accelerating the overall optimization process. A straightforward implementation of aggressive model growth is to reconstruct dense Gaussians from a predefined number of sampled depth points. This approach is similar to depth reinitialization described in Sec. \ref{sec:Densification}, but applied at an early iteration (\eg, 2K steps). While this naive design can considerably accelerate optimization, it suffers from limited scalability to complex or diverse scenes due to its reliance on a fixed sampling hyperparameter. Moreover, it often triggers a sudden spike in the number of Gaussians (\eg, increasing from 0.5M to 4M for \textit{bicycle}), which can destabilize optimization and degrade rendering quality. Therefore, we retain depth reinitialization as the core component but enhance it by incorporating critical Gaussian identification and aggressive Gaussian clone, resulting in our proposed aggressive model growth.

\noindent\textbf{Critical Gaussian Identification.} At the early stage of optimization, the Gaussian representation initialized from sparse SfM points is assumed to be in an under-reconstructed state. Rather than densifying all Gaussians indiscriminately, which may occur when reducing the densification threshold, we aim to selectively identify critical Gaussians that are more likely to lie on the object surface. This targeted component avoids redundancy and promotes structural fidelity. To identify such Gaussians, we derive a reconstruction significance criterion through an approximation of the alpha blending:
\begin{equation}
    \bc(\bx) = \sum_{i=1}^N w_i \bc_i \approx w_{i_{max}} \bc_{i_{max}},
\end{equation}
where $i_{max}=\arg\max\limits_{i} w_i$, and the critical Gaussians $\cG_{i_{max}}$ are those providing the maximum blending weight $w_{i_{max}}$. 
To ensure a minimum contribution of each selected Gaussian, we apply a predefined threshold via the quantile function \cite{li2023compressing} using the 0.99 quantile to filter out a few less significant Gaussians from $\cG_{i_{max}}$, resulting in the final set $\cG^{crit}$.

\noindent\textbf{Aggressive Gaussian Clone.} According to \cite{bulo2024revising, kheradmand20243d}, densifying Gaussians while preserving opacity introduces a bias that implicitly amplifies the influence of densified Gaussians, thereby disrupting the optimization process. This issue becomes more significant in our aggressive model growth (\eg, approximately 5$\%$ or fewer Gaussians are selected at each densification iteration in the progressive strategy, compared to 30$\%$ in our case). To mitigate this problem, we smooth the densification operation in accordance with the state transition method in \cite{kheradmand20243d}. 
Specifically, we adopt Gaussian clone to densify all critical Gaussians, and set the number of cloned Gaussians to 2, which simplifies the computation of the Gaussian center $\bp^{\text {new }} = \bp^{\text {old }}$, opacity $\alpha^{\text {new }}=1-\sqrt{1-\alpha^{\text {old }}}$ and covariance matrix $\bSigma^{\text {new }}=\left(\alpha^{\text {old }}\right)^2 \left( 2\alpha^{\text {new }} - \frac{\left(\alpha^{\text {new }}\right)^{2}}{\sqrt{2}} \right)^{-2} \bSigma^{\text {old }}$.

We further integrate the aforementioned components into the densification stage and ensure compatibility with depth reinitialization. Specifically, the structure-aware distribution scheme is preserved, with critical Gaussian identification and aggressive Gaussian clone applied every 250 iterations, beginning at 500 iterations. Depth reinitialization is conducted at 2K iterations, with the number of sampled depth points adjusted to match the current number of Gaussians. We shorten the total densification period to 3K iterations, with overall optimization limited to 18K steps. Figure \ref{fig:cd_psnr} illustrates the optimization progress of \textit{bicycle} under both progressive and aggressive growth patterns. 
The proposed aggressive model growth schedule significantly increases the number of Gaussians, thereby accelerating early structure emergence, as evidenced by both the geometric similarity of the resulting Gaussians and the corresponding rendering quality.

\subsection{{Occluded Gaussian Culling}}

\begin{figure}[t]
	\centering
	\includegraphics[width=1\linewidth]{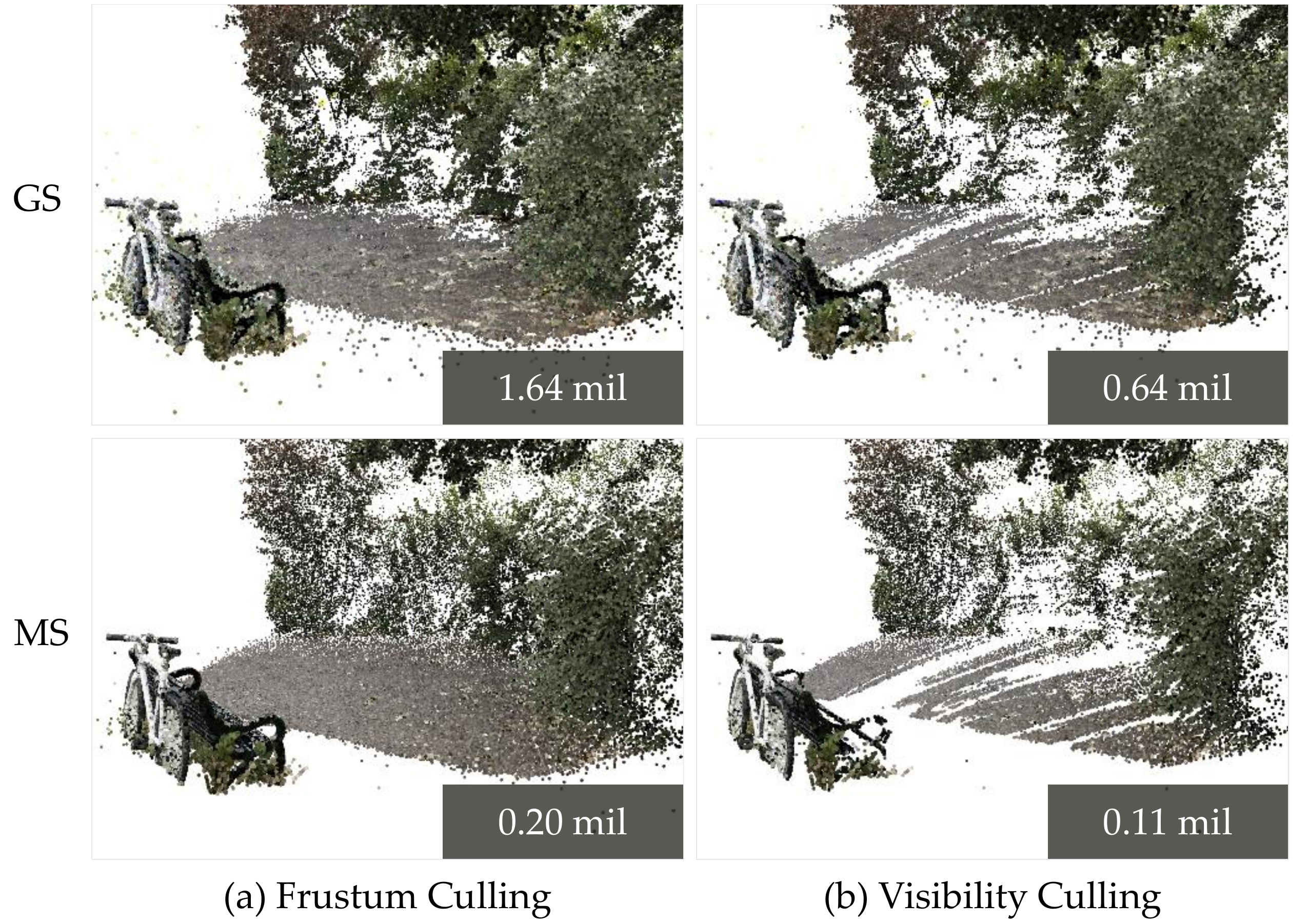}

    \caption{The visualization shows the distribution of Gaussians after applying two distinct culling mechanisms: (a) frustum culling and (b) visibility culling, for both Gaussian Splatting (GS) and Mini-Splatting (MS). Each image highlights the effect of these mechanisms in reducing unnecessary Gaussians, with the respective number of retained Gaussians clearly indicated.}
	\label{fig:culling}
\end{figure}

To further emphasize geometric discrimination during optimization, we incorporate occluded Gaussian culling into the 3DGS framework. The vanilla 3DGS \cite{kerbl20233d} already employs a modified frustum culling, which preserves Gaussians associated with touched tiles during preprocessing. The effects of this culling on typical 3DGS and Mini-Splatting models are illustrated in Fig.~\ref{fig:culling} (a). In this section, we first perform visibility identification and subsequently integrate view-dependent culling into the overall optimization framework to reduce the computational overhead associated with model updates.

\noindent\textbf{{Visibility Identification.}} We explicitly model Gaussian visibility by considering the occlusion relationship between input rays and Gaussian primitives under the point rendering formulation in~\cref{eq:render}. Specifically, for the $k$th training view, the visibility $V_i^k$ of a Gaussian $\cG_i$ is derived from its blending weights in~\cref{eq:render} as follows:
\begin{equation}
    V_i^k = \mathbb I(\sum_{j=1}^J w_{ij}^k>\tau),
\end{equation}
where $J$ denotes the total number of rays intersecting $\cG_i$ at the $k$th view, $\mathbb I$ is the indicator function, and $\tau$ is a predefined visibility threshold, obtained through the quantile function~\cite{li2023compressing} with the keep quantile set at 0.99. As shown in Fig. \ref{fig:culling} (b), the visibility estimated from this formulation acts as a reliable proxy for geometric relevance. Compared to frustum culling, our visibility culling masks a substantially larger set of Gaussians while still preserving critical structural information, thereby reducing computational overhead during optimization.

\noindent\textbf{{View-Dependent Culling.}} For each training view, we compute Gaussian visibility to enable view-dependent culling. In practice, this computation is seamlessly integrated into the optimization pipeline, forming a unified and rearchitected framework. Visibility masks are updated at every training view in conjunction with critical Gaussian identification and redundant Gaussian simplification steps. Occluded Gaussian culling is activated between iterations 500 and 13K, during which all non-visible Gaussians are excluded during the preprocessing stage of rasterization. By focusing computation solely on Gaussians that contribute to the visible scene, this strategy avoids unnecessary updates and reduces training overhead, thus accelerating optimization progress.


\begin{algorithm}[tb]
\caption{{The overall pipeline of MSv2}}
\label{msv2}
\begin{algorithmic}[1]
\State Gaussians $\gets$ SfM Points \Comment{Initialization}
\State $i \gets 0$ \Comment{Iteration Count}
\While{not converged}
    \State OptimizationWithCulling() \Comment{Optimization with  Our Occluded Gaussian Culling}
    \If{$i < $ DensificationIteration}
        \If{IsProgressiveIteration($i$)}
           \State BlurSplit() \Comment{Our Blur Split}
           \State SplitAndClone() \Comment{Split and Clone of 3DGS}
        \EndIf
        \If{IsAggressiveIteration($i$)}
           \State Identification() \Comment{Our Critical Gaussian Identification}
           \State AggressiveClone() \Comment{Our Aggressive Gaussian Clone}
        \EndIf
        \If{IsReinitIteration($i$)}
           \State DepthReinit() \Comment{Our Depth Reinitialization}
        \EndIf
    \Else
        \If{$i=$ SimplificationIteration1}
            \State Intersection() \Comment{Our Intersection Preserving}
            \State Sampling() \Comment{Our Primitive Sampling}
        \EndIf     
        \If{$i=$ SimplificationIteration2}
            \State Intersection() \Comment{Our Intersection Preserving}
            \State Pruning()  \Comment{Directly Prune a Few Gaussians}
        \EndIf      
    \EndIf
    \State $i \gets i+1$
\EndWhile
\end{algorithmic}
\end{algorithm}

\section{Experiments}
\label{Sec:Experiments}


\begin{table*}[htb]
    
    \renewcommand{\tabcolsep}{2pt}
    \centering
    \caption{Quantitative evaluation of Mini-Splatting, Mini-Splatting2 and other methods, categorizing algorithms into sparse and dense Gaussian models. Representative NeRF-based methods are also included as performance references.}
    \label{tab:quant}    
    \resizebox{1\linewidth}{!}{
    \begin{tabular}{@{}l@{\,\,}|ccccc|ccccc|ccccc}
    Dataset & \multicolumn{5}{c|}{Mip-NeRF 360} & \multicolumn{5}{c|}{Tanks\&Temples} & \multicolumn{5}{c}{Deep Blending}  \\
    
  Method $|$ Metric & SSIM $\uparrow$ & PSNR $\uparrow$ & LPIPS $\downarrow$ & Train $\downarrow$ & Num   & SSIM $\uparrow$ & PSNR $\uparrow$ & LPIPS $\downarrow$ & Train $\downarrow$ & Num   & SSIM $\uparrow$ & PSNR $\uparrow$ & LPIPS $\downarrow$ & Train $\downarrow$ & Num \\\hline

mip-NeRF 360~\cite{barron2022mipnerf360}  &                      0.792 &  \cellcolor{tabthird}27.69 &                      0.237 &                       -  &  -  &                 
     0.759 &                      22.22 &                      0.257 &                    
   -  &  -  &                      0.901 &                      29.40 &                   
   0.245 &                       -  &  -  \\ 
Zip-NeRF~\cite{barron2023zip}             & \cellcolor{tabsecond}0.828 &  \cellcolor{tabfirst}28.54 &  \cellcolor{tabthird}0.189 &                       -  &  -  &                 
      -  &                       -  &                       -  &                       -  &  -  &                       -  &           
            -  &                       -  &                       -  &  -  \\ \hline
3DGS~\cite{kerbl20233d}                   &                      0.815 &                  
    27.47 &                      0.216 &                      27min6s & 3.35 &            
          0.848 &                      23.66 &                      0.176 &               
       15min6s & 1.84 &                      
0.904 &                      29.54 &         
             0.244 &                      24min51s & 2.82 \\
3DGS-accel~\cite{taming3dgs} &                      0.811 &                  
    27.38 &                      0.225 &                      10min2s & 2.39 &            
          0.849 &                      23.61 &                      0.175 &               
       6min53s & 1.52 &                      
0.907 &                      29.69 &         
             0.246 &                      8min & 2.31 \\
Taming-L~\cite{taming3dgs}                &                      0.823 & \cellcolor{tabsecond}27.84 &                      0.208 &                      15min44s & 3.22 &  \cellcolor{tabfirst}0.856 &  \cellcolor{tabfirst}24.14 &  \cellcolor{tabthird}0.168 &              
        9min47s & 1.83 & \cellcolor{tabsecond}0.910 & \cellcolor{tabsecond}30.11 &  \cellcolor{tabthird}0.235 &                      12min47s & 2.80 \\
{3DGS-LM*~\cite{hollein20253dgs}}           &                      0.814 &                  
    27.41 &                      0.221 &                      9min49s &  -  &             
         0.844 &  \cellcolor{tabthird}23.72 &                      0.183 &                
      6min33s &  -  &                      0.902 &                      29.72 &           
           0.249 &                      9min &  -  \\
Mini-Splatting-D                          &  \cellcolor{tabfirst}0.831 &                  
    27.51 &  \cellcolor{tabfirst}0.176 &                      32min50s & 4.69 & \cellcolor{tabsecond}0.853 &                      23.23 &  \cellcolor{tabfirst}0.140 &              
        24min24s & 4.28 &                    
  0.906 &                      29.88 &  \cellcolor{tabfirst}0.211 &                      28min26s & 4.63 \\
MSv2-D                                    &  \cellcolor{tabthird}0.827 &                  
    27.54 & \cellcolor{tabsecond}0.184 &                      8min52s & 3.59 &  \cellcolor{tabthird}0.851 &                      23.38 & \cellcolor{tabsecond}0.151 &               
       5min55s & 2.69 &  \cellcolor{tabthird}0.908 &                      29.74 & \cellcolor{tabsecond}0.223 &                      6min23s & 4.11 \\ \hline
Taming-S~\cite{taming3dgs}                &                      0.795 &                  
    27.25 &                      0.259 &  \cellcolor{tabthird}6min28s & 0.67 &            
          0.835 &  \cellcolor{tabthird}23.72 &                      0.211 &  \cellcolor{tabthird}4min27s & 0.32 &                      
0.904 &                      29.77 &         
             0.271 &  \cellcolor{tabthird}4min35s & 0.29 \\
{DashGaussian~\cite{chen2025dashgaussian}}  &                      0.813 &                  
    27.59 &                      0.229 &                      7min12s & 1.98 &            
          0.850 & \cellcolor{tabsecond}24.05 &                      0.186 &               
       5min14s & 1.15 &                      
0.906 &                      29.76 &         
             0.255 &                      4min51s & 1.64 \\         
{Turbo-GS*~\cite{lu2024turbo}}              &                      0.812 &                  
    27.38 &                      0.210 & \cellcolor{tabsecond}5min16s &  -  &             
         0.841 &                      23.49 &                      0.176 & \cellcolor{tabsecond}4min20s &  -  & \cellcolor{tabsecond}0.910 &  \cellcolor{tabfirst}30.41 &           
           0.239 & \cellcolor{tabsecond}3min14s &  -  \\
Mini-Splatting                            &                      0.822 &                  
    27.32 &                      0.217 &                      20min21s & 0.49 &           
           0.846 &                      23.43 &                      0.180 &              
        12min35s & 0.30 & \cellcolor{tabsecond}0.910 &                      29.98 &                      0.241 &                      17min16s & 0.56 \\
MSv2                                      &                      0.821 &                  
    27.33 &                      0.215 &  \cellcolor{tabfirst}3min34s & 0.62 &            
          0.841 &                      23.14 &                      0.186 &  \cellcolor{tabfirst}2min22s & 0.35 &  \cellcolor{tabfirst}0.912 &  \cellcolor{tabthird}30.08 &         
             0.240 &  \cellcolor{tabfirst}2min45s & 0.65

    \end{tabular}
    }
\end{table*}

We begin by presenting our carefully designed models, and then detail the experimental setup, followed by an evaluation in terms of rendering quality and resource consumption. 
In addition, we demonstrate the capability of our algorithm \revise{for} point cloud reconstruction and conduct ablation studies to assess the effectiveness of the proposed components.

\subsection{Model Variant Design}\label{sec:applications}


Leveraging our Gaussian organization and simplification strategies, we develop two model variants, Mini-Splatting and Mini-Splatting-D, which focus on representation compactness. 
By incorporating our optimization acceleration scheme, we extend this framework to Mini-Splatting2 for fast optimization (Algorithm \ref{msv2}), denoting the accelerated sparse- and dense-Gaussian variants as MSv2 and MSv2-D, respectively.

\noindent\textbf{Mini-Splatting.} This variant aims to support resource-efficient rendering under constrained memory budgets. Following an optimization framework similar to that of 3DGS \cite{kerbl20233d}, we set the total number of optimization steps to 30K. Vanilla densification in 3DGS and our adaptive Gaussian organization are performed up to the 15Kth iteration, followed by our redundant Gaussian simplification at both the 15Kth and 20Kth steps to improve compactness. Notably, we observe that incorporating view-dependent color information offers limited benefits during the densification phase. Therefore, we only activate SH coefficients after simplification at 15K to enhance rendering expressiveness in later stages.

\noindent\textbf{Mini-Splatting-D.} This variant is tailored for quality-prioritized rendering. While it adopts a training pipeline similar to Mini-Splatting, the key difference lies in the omission of the simplification step, allowing the model to retain a denser Gaussian set for improved reconstruction fidelity.

\noindent\textbf{MSv2.} This variant is designed for accelerated Gaussian Splatting optimization. Building upon the Mini-Splatting pipeline, MSv2 replaces the original progressive model growth schedule with the proposed aggressive growth, reducing the densification period to 3K iterations and the total optimization steps to 18K. In addition, we incorporate occluded Gaussian culling into optimization to remove redundant computations, establishing a fast foundation for MSv2.

\noindent\textbf{MSv2-D.} This variant extends MSv2 for quality-prioritized rendering, while maintaining fast optimization. To construct a dense yet informative Gaussian representation, we replace the simplification process in MSv2 with importance-based pruning, where accumulated blending weights are used to evaluate Gaussian importance. This approach preserves a larger number of Gaussians, resulting in higher geometric fidelity and improved rendering quality.

\subsection{Experimental Setup}\label{Sec:Setup}


\noindent\textbf{Datasets.} We conduct experiments on three real-world datasets: Mip-NeRF360~\cite{barron2022mipnerf360}, Tanks\&Temples~\cite{Knapitsch2017}, and Deep Blending~\cite{hedman2018deep}. Following the 3DGS~\cite{kerbl20233d} protocol, all datasets are processed consistently, and rendering quality is assessed using the same metrics: peak signal-to-noise ratio (PSNR), structural similarity (SSIM), and LPIPS~\cite{zhang2018unreasonable}.


\noindent\textbf{Implementation Details.} We implement our Gaussian organization and simplification strategies in PyTorch, and integrate them into the optimization pipeline of 3DGS~\cite{kerbl20233d}. In addition, we extend the Gaussian rasterization module to support the rendering of Gaussian indices and depth points. To support our accelerated model variants, we build upon the improved 3DGS implementation introduced in~\cite{taming3dgs}. This enhanced codebase has been merged into the official 3DGS repository and is referred to as 3DGS-accel in our experiments. During the densification stage, we adopt a low-resolution initialization strategy, starting training at half-resolution (\ie, 0.5$\times$) for both MSv2 and MSv2-D to improve efficiency. All experiments are conducted on a workstation equipped with an Intel Xeon Platinum 8383C CPU and a single NVIDIA RTX 3090 GPU.

\subsection{Experimental Results}\label{Sec:Results}

\noindent{\textbf{Quantitative Results.} As shown in Table~\ref{tab:quant}, we compare our proposed variants, Mini-Splatting, Mini-Splatting-D, MSv2, and MSv2-D, against the baseline 3DGS\cite{kerbl20233d, taming3dgs}, Taming-3DGS\cite{taming3dgs}, 3DGS-LM~\cite{hollein20253dgs}, DashGaussian~\cite{chen2025dashgaussian}, Turbo-GS~\cite{lu2024turbo}, and recent NeRF-based methods~\cite{barron2022mipnerf360, barron2023zip}. 3DGS-LM* and Turbo-GS* indicate results collected directly from their reports. The comparison is conducted in terms of training time and standard visual quality metrics, including PSNR, SSIM, and LPIPS. For clarity, we denote Taming-3DGS with a small number of Gaussians as Taming-S and with an increased number of Gaussians as Taming-L.}

The results demonstrate that, compared to the 3DGS baseline, our dense model, Mini-Splatting-D, achieves superior rendering performance across most metrics, highlighting the significant impact of our Gaussian organization in rendering. Notably, our Mini-Splatting-D even surpasses the state-of-the-art neural rendering algorithm, Zip-NeRF \cite{barron2023zip}, in terms of SSIM and LPIPS on the Mip-NeRF360 dataset. Moreover, Mini-Splatting remains competitive with 3DGS while utilizing only a fraction (7$\times$ fewer) of the Gaussians.

\begin{figure*}[tb]
	\centering
	\includegraphics[width=0.8\linewidth]{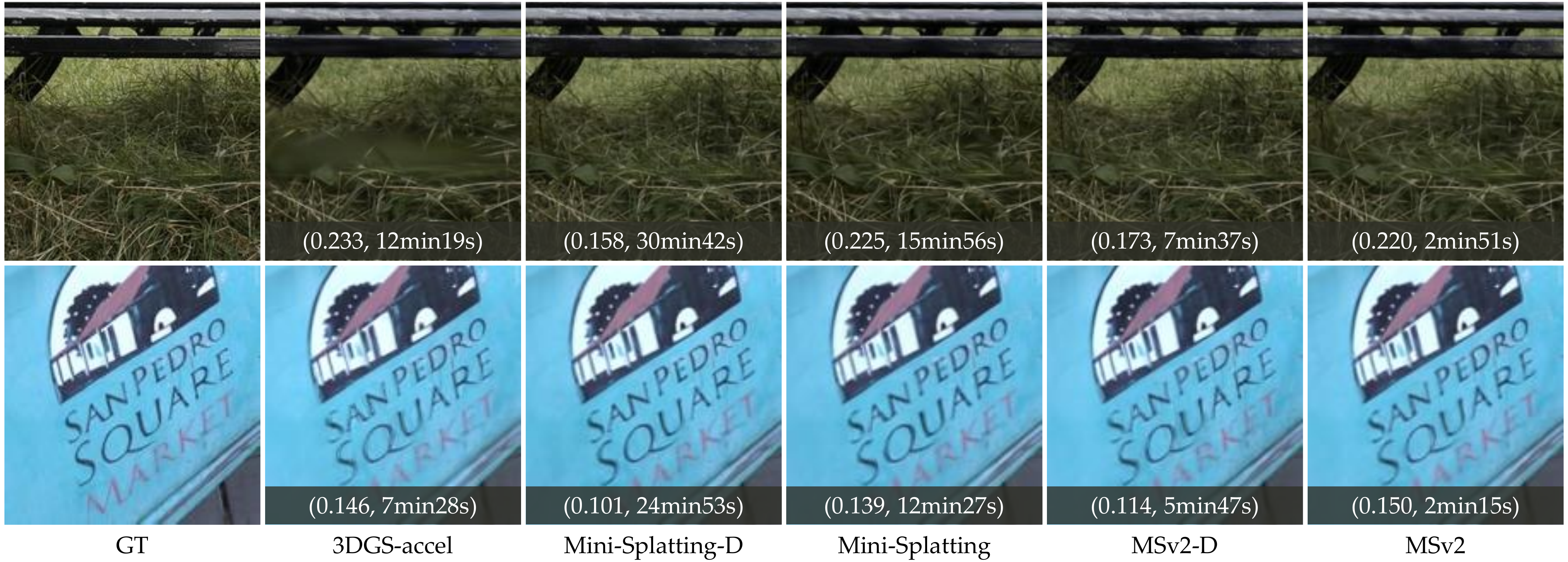}
    \caption{Qualitative results of Mini-Splatting, Mini-Splatting2 and other methods. LPIPS and optimization time are displayed as (LPIPS, Time).}
	\label{fig:qual}
\end{figure*}

For our accelerated model variants, the results indicate that MSv2 achieves rendering quality comparable to both 3DGS and 3DGS-accel while significantly improving optimization speed. Specifically, MSv2 accelerates training by factors of 7.6$\times$ and 2.8$\times$ over the original 3DGS and 3DGS-accel, respectively, while maintaining similar rendering quality. Compared to the recently proposed Taming-S, MSv2 delivers better visual quality with lower training time and a comparable number of Gaussians. Additionally, our dense Gaussian model, MSv2-D, which requires a longer yet reasonable training time, demonstrates superior rendering quality, with only a slight performance drop compared to its dense counterpart, Mini-Splatting-D. Moreover, MSv2-D provides a notable improvement in rendering quality over 3DGS-accel, while remaining within the same optimization time budget.

Notably, our algorithms exhibit a decreased PSNR on the Tanks\&Temples dataset. This can be attributed to the presence of large textureless sky regions in certain scans (\ie, \textit{train}) within Tanks\&Temples. Since our method employs a depth-based strategy (\ie, depth reinitialization) to reinitialize Gaussian models, it struggles in regions lacking reliable depth values. In contrast, our algorithms perform well in other cases, particularly for scans with well-defined depth, as demonstrated in the Mip-NeRF 360 and Deep Blending datasets.

\noindent\textbf{Qualitative Results.} The qualitative results for \textit{bicycle} and \textit{truck} are presented as rendered images in Fig.~\ref{fig:qual}. These results are consistent with the quantitative findings reported in Table \ref{tab:quant}. All of our models, MSv2, MSv2-D, Mini-Splatting, and Mini-Splatting-D, show improved performance over the baseline 3DGS-accel, particularly in the \textit{bicycle} scan, where foreground regions exhibit fewer blurry artifacts. These improvements can be attributed to the adaptive Gaussian organization module, which is incorporated across all Mini-Splatting variants.

Furthermore, it is clear that Mini-Splatting-D and MSv2-D demonstrate superior rendering quality, effectively preserving fine-grained scene details. In contrast, Mini-Splatting and MSv2 maintain competitive visual performance while significantly reducing the number of Gaussians. The slight performance drop observed from Mini-Splatting-D to Mini-Splatting, and from MSv2-D to MSv2, primarily affects background regions. This can be attributed to the simplification strategy, which intentionally trades off marginal visual fidelity for greater compactness and memory efficiency.

In summary, compared to 3DGS-accel, our proposed MSv2 achieves comparable visual quality while substantially reducing both the number of Gaussians and optimization time. Meanwhile, MSv2-D further enhances rendering quality without incurring additional optimization cost. These results collectively highlight the effectiveness of our proposed models in achieving a favorable balance between compact representation and fast optimization of Gaussians, thereby enabling efficient and high-quality 3D scene modeling.

\begin{table}[t]
    \renewcommand{\tabcolsep}{2pt}
    \centering
    \caption{Resource consumption of our algorithms and other approaches on Mip-NeRF360. Num and Peak represent the final and peak number of Gaussians (in millions) during optimization. Time indicates the total time required for optimization, and Mem indicates the peak GPU memory usage (in GB) during optimization.}
    \label{tab:resource} 
    \resizebox{1\linewidth}{!}{
\begin{tabular}{@{}l@{\,\,}|cccc|cccc}
     & \multicolumn{4}{c|}{Mip-NeRF360 (Outdoor)} & \multicolumn{4}{c}{Mip-NeRF360 (Indoor)}\\ \hline
    Method $|$ Metric & Num & Peak & Time & Mem & Num & Peak & Time & Mem  \\ \hline

    3DGS~\cite{kerbl20233d} & 4.86  & 4.86  & 30min8s & 7.45 & 1.46  & 1.46  & 24min41s & 2.75 \\
    3DGS-accel~\cite{taming3dgs} & 3.34  & 3.34  & 11min5s & 5.03 & 1.29  & 1.29  & 10min50s & 2.64 \\ \hline
    Mini-Splatting-D & 5.40  & 5.40  & 31min48s & 7.45 & 3.80  & 3.80  & 40min13s & 5.55 \\
    MSv2-D & 3.58  & 4.60  & 7min46s & 6.60 & 3.57  & 4.86  & 11min & 7.53 \\
    Mini-Splatting & 0.57  & 5.40  & 17min56s & 2.61 & 0.40  & 3.80  & 27min2s & 2.77 \\
    MSv2 & 0.65  & 4.61  & 3min20s & 3.70 & 0.58  & 4.89  & 4min59s & 4.55 

    \end{tabular}%
    }
\end{table}

\noindent\textbf{Resource Consumption.} We also present the resource consumption results of our algorithm, including the final and peak number of Gaussians during optimization, training time, and peak GPU memory usage, as summarized in Table \ref{tab:resource}. Peak memory consumption is measured using \texttt{torch.cuda.max\_memory\_allocated()}. Notably, this experiment is conducted under limited-resource conditions, where source data is loaded into CPU memory, resulting in a slight slowdown in training compared to Table~\ref{tab:quant}.

For outdoor scenes, Mini-Splatting-D, despite utilizing a larger number of Gaussians, exhibits similar training time and memory consumption as 3DGS, thanks to our strategy of constraining SH coefficients during densification. Furthermore, Mini-Splatting, with its sparser Gaussian representation, significantly accelerates training while reducing peak memory usage. MSv2, however, incurs an increase in GPU memory consumption compared to Mini-Splatting due to the precomputed visibility, although this increase remains affordable.

In the case of indoor scenes, the peak number of Gaussians indicates that MSv2 reconstructs more Gaussians than Mini-Splatting and 3DGS during optimization. This increase stems from our aggressive model growth schedule, which allocates additional Gaussians to textureless regions, such as walls and floors, to enhance rendering quality. It is worth noting that the \textit{kitchen} scan, which has the highest GPU memory demand, is reconstructed with 5.2 million Gaussians after densification while maintaining a peak memory consumption of only 5.0GB. These results demonstrate that our algorithm remains feasible for use with low-cost graphics cards.


From this experiment, we take a distinct perspective compared to Taming-3DGS \cite{taming3dgs}. Instead of constraining the peak number of Gaussians, we find that limiting SH coefficients (\ie, 48 out of 59 optimizable Gaussian attributes) is a more practical and effective approach under resource constraints. Although our algorithm is not explicitly designed for memory-constrained scenarios, it remains feasible to train and evaluate our proposed Mini-Splatting on a low-cost graphics card (\eg, a GTX 1060 6G GPU). Furthermore, these results support our aggressive growth and simplification schedule over the non-aggressive alternative, which simply limits the number of Gaussians throughout the optimization process.


\begin{figure}[tb]
	\centering
	\includegraphics[width=1\linewidth]{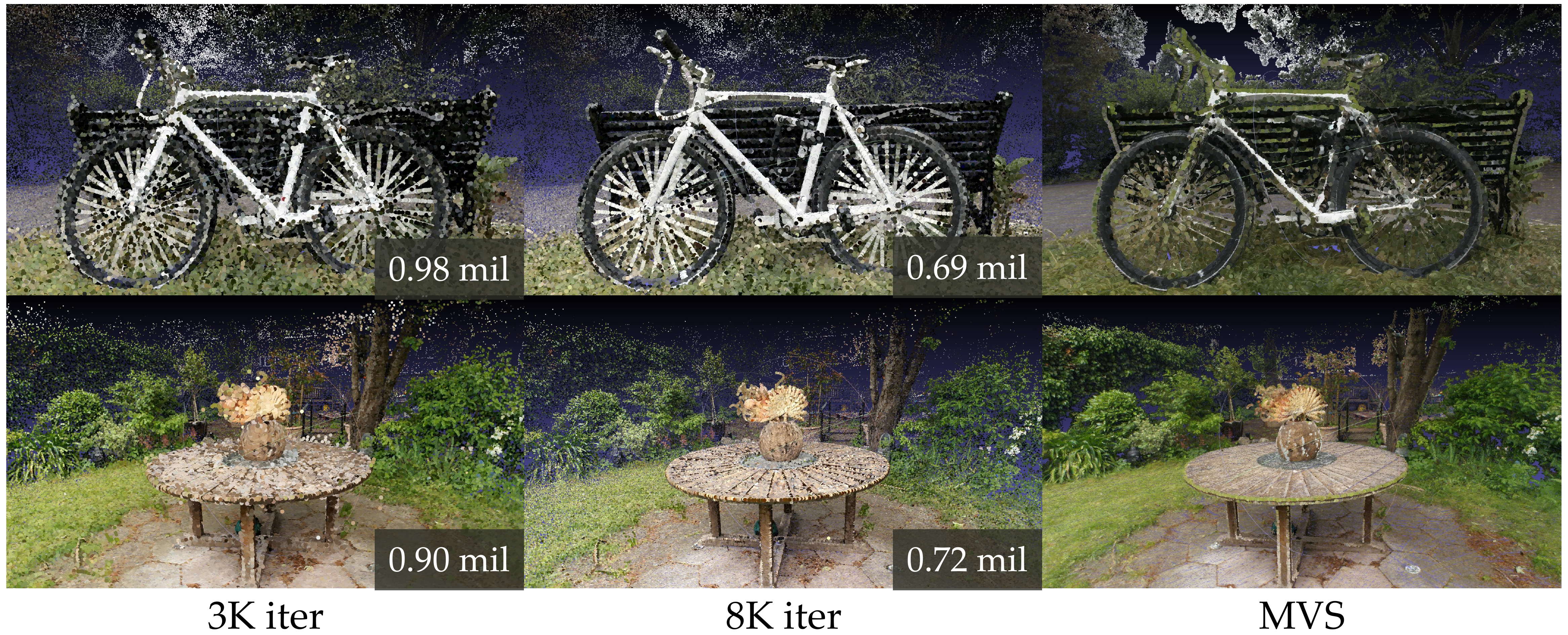}
    \caption{Direct visualization of Gaussian centers from Mini-Splatting2 and MVS points \cite{schoenberger2016mvs} in MeshLab. The comparison illustrates the potential of our method to produce dense and well-structured point clouds, demonstrating its effectiveness for accelerated 3D reconstruction.
    }
	\label{fig:pc_rec}
\end{figure}

\subsection{Point Cloud Reconstruction} 

To substantiate our assumption that Mini-Splatting2 supports dense point cloud reconstruction, we provide a direct and faithful visualization of Gaussian centers and MVS points \cite{schoenberger2016mvs} in MeshLab in Fig. \ref{fig:pc_rec}. After a brief 3K-iteration densification and subsequent simplification (within 1 minute), Mini-Splatting2 effectively reconstructs the coarse scene geometry, albeit with some floaters. Continuing with an additional 5K-iteration optimization and further simplification (taking an additional 1 to 2 minutes), the algorithm produces a high-quality point cloud, approaching the fidelity of the dense MVS points. Although this study primarily focuses on fast optimization for novel view synthesis, the algorithm also demonstrates potential for efficiency-driven geometry reconstruction methods.

\begin{table}[tb]
    
    \renewcommand{\tabcolsep}{6pt}
    \centering
    \caption{Ablation study of adaptive Gaussian organization on the Mip-NeRF360 dataset. Blur split and depth reinitialization (Depth Reinit) are sequentially added to the baseline.
    }
    \label{tab:abl_densification}    
    \resizebox{0.80\linewidth}{!}{
\begin{tabular}{@{}l@{\,\,}|cccc}

        \Xhline{2.0\arrayrulewidth}
      & SSIM $\uparrow$ & PSNR $\uparrow$ & LPIPS $\downarrow$ & Num \\ \hline
    Baseline & 0.815  & 27.47  & 0.216  & 3.35  \\
    + Blur Split & 0.819  & 27.47  & 0.195  & 3.74  \\
    + Depth Reinit & 0.832  & 27.54  & 0.175  & 4.32  \\
        \Xhline{2.0\arrayrulewidth}
    \end{tabular}%
    }
\end{table}



\begin{table}[tb]
    
    \renewcommand{\tabcolsep}{6pt}
    \centering
    \caption{Comparison of depth variants for depth reinitialization.
    }
    \label{tab:abl_depth}    
    \resizebox{0.71\linewidth}{!}{
      \begin{tabular}{@{}l@{\,\,}|cccc}
        \Xhline{2.0\arrayrulewidth}
      & SSIM $\uparrow$ & PSNR $\uparrow$ & LPIPS $\downarrow$ & Num \\ \hline
    Blending & 0.513  & 17.67  & 0.475  & 4.01  \\
    Center & 0.832  & 27.57  & 0.176  & 4.70  \\
    Mid   & 0.832  & 27.54  & 0.175  & 4.69  \\
        \Xhline{2.0\arrayrulewidth}
    \end{tabular}%
    }
\end{table}



\begin{figure}[tb]
	\centering
	\includegraphics[width=0.69\linewidth]{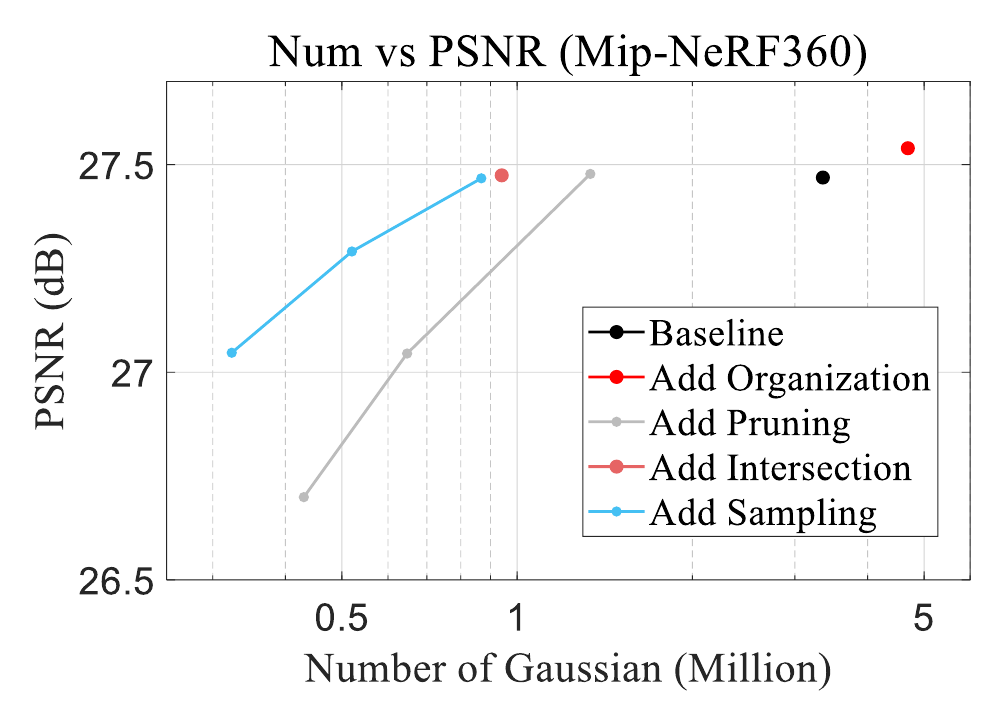}
    \caption{Ablations of Gaussian simplification on the Mip-NeRF360 dataset. Components are sequentially added to the baseline implementation.
    }
	\label{fig:abl_simplification}
\end{figure}

\subsection{Ablation Study on Structure-Aware Distribution Scheme}

\noindent\textbf{Adaptive Gaussian Organization.} We begin by presenting the ablation study of our Gaussian organization strategy (blur split and depth reinitialization) in Table~\ref{tab:abl_densification}. Starting from the 3DGS baseline, we incrementally introduce the blur split and depth reinitialization steps. The results indicate a consistent improvement in rendering quality as the number of Gaussians increases. 
Notably, we observe a strong correlation between the number of Gaussians and the LPIPS metric, suggesting that a denser Gaussian representation \revise{demonstrates improved} perceptual quality as measured by human-centric metrics.

We further extend our analysis by comparing the blending depth with depth values obtained from the Gaussian center and the midpoint, as summarized in Table~\ref{tab:abl_depth}. As discussed in Sec.~\ref{sec:Densification}, reinitializing Gaussians using blending depth may degrade rendering quality due to the presence of noisy points introduced by depth artifacts. In contrast, depth reinitialization based on the other two strategies yields improved reconstruction results. While using the Gaussian center for reinitialization produces results comparable to those obtained from the midpoint, the latter demonstrates superior performance in dense point cloud reconstruction, as illustrated in Fig.~\ref{fig:merged_depth}, and holds potential for extension to geometry-related tasks.

\noindent\textbf{Redundant Gaussian Simplification.} The ablation study of our Gaussian simplification on the Mip-NeRF360 dataset is illustrated in Fig. \ref{fig:abl_simplification}. Beginning with the 3DGS baseline, we first include our Gaussian Organization to obtain our Mini-Splatting-D (\ie, Add Organization). Next, we integrate direct importance pruning into Mini-Splatting-D and adjust the pruning ratio to establish the baseline curve, denoted as Add Pruning. Notably, this baseline implementation represents the combination of Mini-Splatting-D and direct pruning, surpassing the performance of 3DGS with pruning.

To underscore the effectiveness of our intersection preserving and primitive sampling components, we sequentially add intersection preserving and then primitive sampling to Mini-Splatting-D (denoted as Add Intersection and Add Sampling). The results clearly demonstrate that our simplification strategy outperforms direct pruning, achieving approximately a 50\% reduction in the total number of Gaussians while maintaining comparable rendering quality. 
This outcome highlights the effectiveness of the proposed redundant Gaussian simplification strategy and further underscores the benefits of eliminating irregularity in the Gaussian representation.

\begin{table}[tb]
    
    \renewcommand{\tabcolsep}{5pt}
    \centering
    \caption{Ablation study on the acceleration scheme on the Mip-NeRF360 dataset. Aggressive model growth (Agg Growth) and occluded Gaussian culling (Occ Culling) are sequentially added to the baseline.
    }
    \label{tab:component}    
    \resizebox{0.98\linewidth}{!}{
      \begin{tabular}{@{}l@{\,\,}|ccccc}

        \Xhline{2.0\arrayrulewidth}
      & SSIM $\uparrow$ & PSNR $\uparrow$ & LPIPS $\downarrow$ & Train $\downarrow$ & Num \\ \hline

    3DGS-accel &  0.811 &  27.38  &  0.225 & 10min2s & 2.39 \\
    Mini-Splatting & 0.822 &   27.32  &  0.217 & 20min21s & 0.49 \\
    Mini-Splatting* & 0.823 & 27.23 & 0.214 & 7min59s & 0.56 \\ \hline
    + Agg Growth & 0.822 & 27.43 & 0.212 & 4min4s & 0.64 \\
    + Occ Culling & 0.821 & 27.33 & 0.215 & 3min34s & 0.62 \\
        \Xhline{2.0\arrayrulewidth}
    \end{tabular}
    }
\end{table}

    



\begin{table}[tb]
    
    \renewcommand{\tabcolsep}{5pt}
    \centering
    \caption{Comparison of densification implementations that support aggressive model growth.
    }
    \label{tab:densification}    
    \resizebox{0.98\linewidth}{!}{
      \begin{tabular}{@{}l@{\,\,}|ccccc}

        \Xhline{2.0\arrayrulewidth}
      & SSIM $\uparrow$ & PSNR $\uparrow$ & LPIPS $\downarrow$ & Train $\downarrow$ & Num \\ \hline
    Mini-Splatting* & 0.823 & 27.23 & 0.214 & 7min59s & 0.56 \\
    MSv2 (Ours) & 0.821 & 27.33 & 0.215 & 3min34s & 0.62 \\ \hline
    w 3K Dens & 0.776 & 26.49 & 0.285 & 2min23s & 0.25 \\
    w Low Thres & 0.799 & 26.58 & 0.237 & 3min43s & 0.59 \\
    w Agg-Depth & 0.817 & 27.12 & 0.215 & 3min14s & 0.62 \\
    w Agg-Full & 0.821 & 27.33 & 0.215 & 3min34s & 0.62 \\
        \Xhline{2.0\arrayrulewidth}
    \end{tabular}
    }
\end{table}

\subsection{Ablation Study on Region-Prioritized Optimization Scheme}

\noindent\textbf{Effectiveness of Acceleration Scheme.} To isolate the impact of Mini-Splatting and 3DGS-accel on optimization speed, we conduct an ablation study on the Mip-NeRF360 dataset, evaluating the effectiveness of each improvement in Table \ref{tab:component}. Starting with the baseline, Mini-Splatting*, which combines Mini-Splatting and 3DGS-accel, we incrementally add aggressive model growth (Agg Growth) and occluded Gaussian culling (Occ Culling). 
Comparison results show that \revise{our aggressive model growth halves the training time}, while occluded Gaussian culling further accelerates optimization by an additional half-minute.


\noindent\textbf{Aggressive Model Growth.} Table~\ref{tab:densification} presents the results of our ablation study on aggressive model growth. Starting with the MSv2 model, we include several densification variants to support our aggressive model growth: 3K Dens (a shortened progressive densification over 3K iterations), Low Thres (a lower gradient threshold for densification), Agg-Depth (a predefined large number of points for depth initialization), and Agg-Full (our complete aggressive model growth). All aggressive growth strategies yield substantial speedups over the Mini-Splatting* baseline. Among them, Agg Growth-Full not only adaptively regulates the Gaussian count but also improves rendering quality. These results demonstrate the effectiveness of our aggressive model growth strategy in terms of both scalability and visual fidelity.

\begin{table}[tb]
    
    \renewcommand{\tabcolsep}{5pt}
    \centering
    \caption{Comparison of significance criteria for critical Gaussian identification.
    }
    \label{tab:appendix_iden}    
    \resizebox{0.86\linewidth}{!}{
      \begin{tabular}{@{}l@{\,\,}|ccccc}

        \Xhline{2.0\arrayrulewidth}
      & SSIM $\uparrow$ & PSNR $\uparrow$ & LPIPS $\downarrow$ & Train $\downarrow$ & Num \\ \hline
    Random & 0.819  & 27.27 & 0.218 & 3min47s & 0.54 \\
    Opacity & 0.820  & 27.28 & 0.215 & 3min49s & 0.62 \\
    $\sum_{i} w_i$  & 0.820  & 27.34 & 0.215 & 3min46s & 0.62 \\
    $w_{i_{max}}$  & 0.820  & 27.36 & 0.215 & 3min36s & 0.62 \\
        \Xhline{2.0\arrayrulewidth}
    \end{tabular}
    }
    \vspace{-0.3cm}
\end{table}

\begin{table}[tb]
    
    \renewcommand{\tabcolsep}{5pt}
    \centering
    \caption{Comparison of implementation variants for aggressive Gaussian clone.}
    \label{tab:appendix_clone}    
    \resizebox{0.95\linewidth}{!}{
      \begin{tabular}{@{}l@{\,\,}|ccccc}

        \Xhline{2.0\arrayrulewidth}
      & SSIM $\uparrow$ & PSNR $\uparrow$ & LPIPS $\downarrow$ & Train $\downarrow$ & Num \\ \hline
    Vanilla & 0.818  & 27.04 & 0.213 & 3min28s & 0.68 \\
    Add $\alpha^{\text {new}}$ \cite{bulo2024revising} & 0.820  & 27.37 & 0.217 & 3min34s & 0.59 \\
    Add $\bSigma^{\text {new}}$ \cite{kheradmand20243d} & 0.820  & 27.36 & 0.215 & 3min36s & 0.61 \\
        \Xhline{2.0\arrayrulewidth}
    \end{tabular}
    }
    \vspace{-0.3cm}
\end{table}

\noindent\textbf{Critical Gaussian Identification.} Table~\ref{tab:appendix_iden} presents the results of our ablation study on critical Gaussian identification for the Mip-NeRF360 dataset. In this study, we compare different significance criteria for identifying critical Gaussians: random selection (Random), opacity (Opacity), accumulated blending weights ($\sum_{i} w_i$) and maximum blending weights ($w_{i_{max}}$). For each criterion, the predefined threshold is adjusted to maintain a consistent number of Gaussians. The results demonstrate that weight-based criteria, specifically $\sum_{i} w_i$ and $w_{i_{max}}$, consistently perform better than other choices. Among these, maximum blending weights yield the best performance for our aggressive model growth approach.

While incorporating image-space information (\eg, 2D gradient and error) and inherent Gaussian attributes (\eg, opacity and scales) might further refine the identification criterion, such methods would introduce case-dependent and hand-crafted heuristics. 
To maintain simplicity and robustness, we adopt the maximum blending weights, $w_{i_{max}}$, as our \revise{significance criterion} for critical Gaussian identification.

\noindent\textbf{Aggressive Gaussian Clone.} Table~\ref{tab:appendix_clone} presents the results of our ablation study on aggressive Gaussian clone for the Mip-NeRF360 dataset. This study specifically compares the detailed densification operations. In Table~\ref{tab:appendix_clone}, `Vanilla' refers to the original Gaussian clone and split in 3DGS \cite{kerbl20233d}. As noted in Sec. \ref{sec:Aggressive_Gaussian_Densification}, this operation introduces a bias that implicitly amplifies the influence of densified Gaussians, thereby disrupting the optimization process \cite{bulo2024revising}. To address this issue, our aggressive Gaussian clone only performs the clone operation and replaces the opacity of the resulting Gaussians, following \cite{bulo2024revising}, denoted as `Add $\alpha^{\text {new}}$'. Additionally, we incorporate the replacement of the covariance matrix as described in \cite{kheradmand20243d}, which is indicated as `Add $\bSigma^{\text {new}}$'. As shown in Table~\ref{tab:appendix_clone}, for our aggressive model growth, both modified operations achieve similar rendering quality, surpassing the performance of the vanilla operation. 
For our aggressive Gaussian clone, we adopt the clone formulation proposed in \cite{kheradmand20243d}, as it \revise{supports} a more general densification \revise{operation} (\ie, clone a single Gaussian into a predefined number of Gaussians).

\begin{table}[tb]
    
    \renewcommand{\tabcolsep}{5pt}
    \centering
    \caption{
    {Ablation study on the number of optimization iterations for aggressive model growth.}
    }
    \label{tab:30k}    
    \resizebox{0.95\linewidth}{!}{
      \begin{tabular}{@{}l@{\,\,}|ccccc}

        \Xhline{2.0\arrayrulewidth}
      & SSIM $\uparrow$ & PSNR $\uparrow$ & LPIPS $\downarrow$ & Train $\downarrow$ & Num \\ \hline

    3DGS-accel & 0.812 & 27.39 & 0.223 & 8min57s & 2.39 \\
    Mini-Splatting* & 0.823 & 27.23 & 0.214 & 7min59s & 0.56 \\
    MSv2-30K & 0.823 & 27.56 & 0.211 & 5min28s & 0.62 \\ \hline
    MSv2-18K & 0.821 & 27.33 & 0.215 & 3min34s & 0.62 \\
        \Xhline{2.0\arrayrulewidth}
    \end{tabular}
    }
\end{table}

\begin{table}[tb]
    
    \renewcommand{\tabcolsep}{5pt}
    \centering
    \caption{{Ablation on culling iterations. MSv2 without culling is denoted as `w/o', and with culling terminated at $N$K iterations as $N$K.}
    }
    \label{tab:culling_iter}    
    \resizebox{0.81\linewidth}{!}{
      \begin{tabular}{@{}l@{\,\,}|ccccc}

        \Xhline{2.0\arrayrulewidth}
      & SSIM $\uparrow$ & PSNR $\uparrow$ & LPIPS $\downarrow$ & Train $\downarrow$ & Num \\ \hline

    w/o & 0.822 & 27.43 & 0.212 & 4min4s & 0.64 \\
    3K    & 0.821  & 27.40  & 0.214 & 3min52s & 0.62 \\
    8K    & 0.821  & 27.38  & 0.214 & 3min39s & 0.62 \\
    13K   & 0.821  & 27.37  & 0.215 & 3min32s & 0.62 \\
    15K   & 0.820  & 27.34  & 0.215 & 3min29s & 0.62 \\
    17K   & 0.820  & 27.31  & 0.215 & 3min26s & 0.62 \\
    18K   & 0.815  & 26.80  & 0.217 & 3min26s & 0.62 \\
        \Xhline{2.0\arrayrulewidth}
    \end{tabular}
    }
\end{table}

\begin{figure*}[tb]
	\centering
	\includegraphics[width=0.87\linewidth]{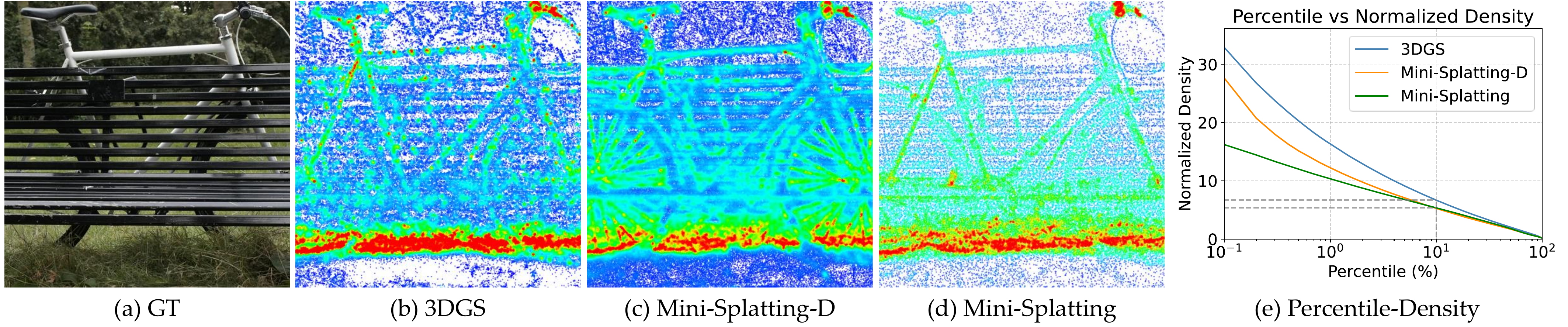}
    \caption{
    {Clustering behavior of Gaussians. (a) GT image. (b-d) Projected Gaussian centers from 3DGS, Mini-Splatting-D, and Mini-Splatting, colored by normalized density. (e) Percentile-density curves showing the fraction of Gaussians above threshold $y$.}
    }
	\label{fig:density}
\end{figure*}

\noindent{\textbf{Ablation on Optimization Iterations.} To validate the effectiveness of aggressive model growth, we conduct an ablation study that isolates the impact of optimization iterations.
As summarized in Table~\ref{tab:30k}, MSv2-30K denotes our model optimized with 30K iterations, while MSv2-18K refers to the reported model using 18K iterations. With the same number of optimization iterations as 3DGS-accel and Mini-Splatting*, MSv2-30K still achieves a substantial improvement in optimization speed (39\%, 32\%) along with a consistent performance gain. Notably, MSv2-18K delivers further speed-up while maintaining comparable SSIM and LPIPS to MSv2-30K. This controlled comparison on optimization iterations highlights the effectiveness of our aggressive model growth.}

\noindent{\textbf{Occluded Gaussian Culling.} To further validate the effectiveness of occluded Gaussian culling, we vary the number of culling iterations as shown in Table~\ref{tab:culling_iter}. MSv2 without occluded Gaussian culling is denoted as `w/o', and MSv2 with culling terminated after $N$K iterations is denoted as $N$K. As the number of culling iterations increases, the training time decreases, while rendering quality experiences a slight drop. Notably, MSv2 with culling applied throughout all iterations (\ie, `18K') exhibits relatively low rendering quality. This is because visibility information is only available for training views and not for unseen test views, leading to overfitting when culling is applied excessively. Nevertheless, compared to MSv2 without culling (\ie, `w/o'), the reported MSv2 model (\ie, `13K') achieves a notable speed-up of approximately half a minute with only a minor performance drop, demonstrating the effectiveness of our occluded Gaussian culling.}

\begin{figure*}[tb]
	\centering
	\includegraphics[width=0.87\linewidth]{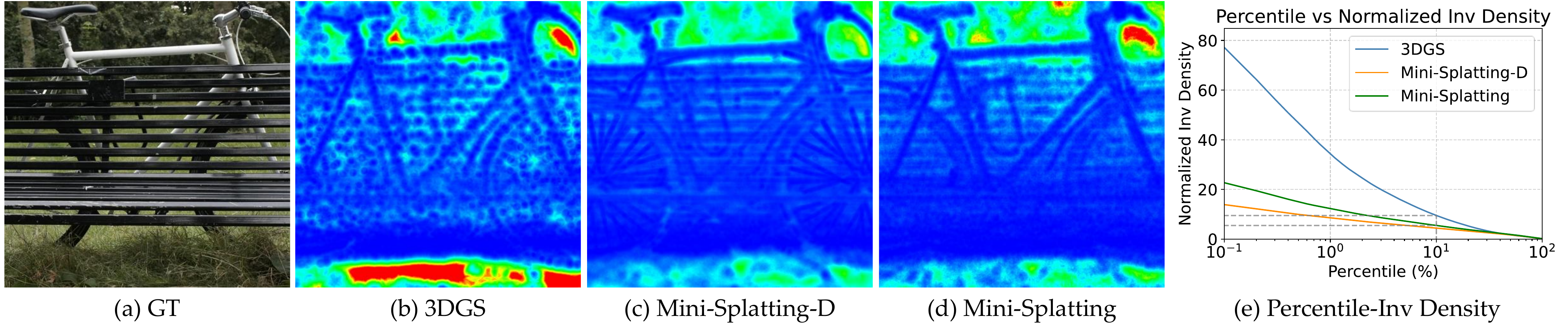}
    \caption{
    {Image-space coverage of Gaussians. (a) GT image. (b-d) Sampled pixels from 3DGS, Mini-Splatting-D, and Mini-Splatting, colored by normalized inverse density. (e) Percentile-inverse density curves showing the fraction of image area above threshold $y$.}
    }
	\label{fig:inv_density}
\end{figure*}

{
\section{Discussion on Spatial Distribution of 3D Gaussians}
\label{Sec:Discussion}

\subsection{Improvement of Irregular Distribution}

To demonstrate the improvement of our method in the spatial distribution of Gaussians, we provide a detailed comparison between vanilla 3DGS and Mini-Splatting with respect to the `overlapping' and `under-reconstruction' issues, both of which indicate irregular distributions. The `overlapping' issue is characterized by the clustering behavior of Gaussians and can be quantified using the density of clustered Gaussians (\ie, the normalized density of each projected Gaussian). 
In contrast, the `under-reconstruction' issue is reflected in the image-space coverage of Gaussians and can be evaluated by the proportion of the image area covered by Gaussians (\ie, the normalized inverse density across the image space).

\noindent\textbf{Validation of Overlapping.} We first examine `overlapping' by analyzing the clustering behavior of Gaussians.
To quantify the density of clustered Gaussians during rendering and optimization, we project the Gaussian centers into the image space, and compute the projected Gaussian density $D$ for each Gaussian using KNN search: $D = \frac{N_G}{L_{N_G}}$, where $N_G$ denotes the number of neighboring Gaussians for this Gaussian and $L_{N_G}$ is the corresponding distance obtained from KNN. For a fair comparison across models, we normalize the density $D$ by the total number of Gaussians $N$ in the corresponding view, yielding the normalized density $D_{norm}=\frac{D}{N}$. The resulting distributions are visualized as heatmaps, where values range from low (blue) to high (red).

As illustrated in Fig. \ref{fig:density} (b) to (d), our model variants, Mini-Splatting-D and Mini-Splatting, exhibit smoother density transitions on the foreground objects compared with the baseline 3DGS, which displays more pronounced clustering. Furthermore, in Fig. \ref{fig:density} (e), we present percentile-density curves, where each point indicates that the normalized densities of the top $x\%$ Gaussians exceed a threshold $y$. These statistics clearly show that Mini-Splatting consistently achieves lower normalized densities than 3DGS, particularly within the 0\% to 20\% range, suggesting reduced Gaussian clustering.

\begin{table}[t]
    \renewcommand{\tabcolsep}{3pt}
    \centering
    \caption{
    {Normalized density of the top 1\%, 10\%, and 20\% Gaussians, and normalized inverse density of the top 1\%, 10\%, and 20\% areas on the Mip-NeRF360 dataset.}
    }
    \label{tab:dist}
    \resizebox{0.95\linewidth}{!}{
\begin{tabular}{@{}l@{\,\,}|ccc|ccc}
        \Xhline{2.0\arrayrulewidth}
    & \multicolumn{3}{c|}{Normalized Density} & \multicolumn{3}{c}{Normalized Inverse Density}\\ \hline
    Method & 1\%   & 10\%  & 20\%  & 1\%   & 10\%  & 20\% \\\hline
    3DGS  & 21.35  & 7.42  & 4.85  & 285.08  & 95.40  & 34.84  \\
    Mini-Splatting-D & 11.09  & 4.31  & 2.96  & 10.87  & 5.24  & 3.63  \\
    Mini-Splatting & 11.21  & 5.10  & 3.61  & 69.10  & 20.63  & 9.18  \\

        \Xhline{2.0\arrayrulewidth}
    \end{tabular}%
    }
\end{table}

\begin{figure*}[tb]
	\centering
	\includegraphics[width=0.87\linewidth]{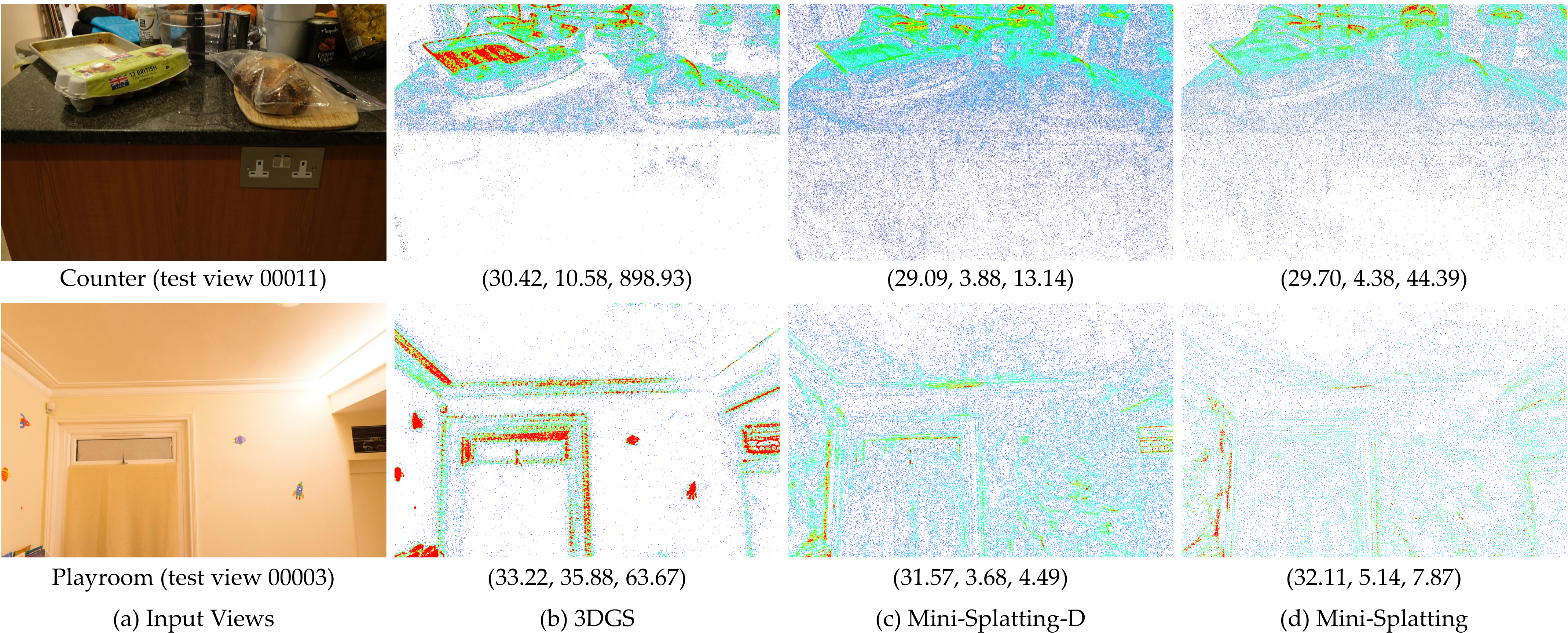}
    \caption{
    {Under-reconstruction case. (a) Inputs of \textit{counter} and \textit{playroom}. (b-d) Projected Gaussian centers from 3DGS, Mini-Splatting-D, and Mini-Splatting. PSNR, normalized density, and normalized inverse density at top 10\% Gaussians/areas are reported as (PSNR, normalized density, normalized inverse density).}
    }
	\label{fig:fail}
\end{figure*}

\noindent\textbf{Validation of Under-Reconstruction.} We then investigate the `under-reconstruction' issue by analyzing the image-space coverage of Gaussians. To quantify this coverage during rendering and optimization, we uniformly sample pixels in the image space and compute the Gaussian density at each pixel $D^{pixel}$ using KNN search, considering the number of neighboring Gaussians and their distances to the pixel. 
We then define the normalized inverse density as $D_{norm}^{inv} = \frac{1}{D_{norm}^{pixel}} = \frac{N}{D^{pixel}}$ and visualize it as heatmaps, where higher values indicate regions of sparse Gaussian coverage.

As shown in Fig. \ref{fig:inv_density} (b) to (d), compared with 3DGS, our variants Mini-Splatting-D and Mini-Splatting significantly reduce the inverse density in foreground regions, thereby increasing local density in areas with insufficient coverage. In addition, Figure \ref{fig:inv_density} (e) reports percentile-inverse density curves, where each point indicates that the normalized inverse densities of the top $x\%$ image area exceed a threshold $y$. These results demonstrate that Mini-Splatting consistently achieves lower normalized inverse densities than 3DGS across the image plane, particularly within the 0\% to 20\% range, indicating improved scene coverage.

\noindent\textbf{Statistics across Scenes.} As shown in Table~\ref{tab:dist}, we further present the statistical analysis of normalized density and inverse density on the Mip-NeRF360 dataset. Specifically, we report the normalized density for the top 1\%, 10\%, and 20\% Gaussians, as well as the normalized inverse density for the top 1\%, 10\%, and 20\% of the image area. The results indicate that, compared to 3DGS (21.35 and 285.08), Mini-Splatting achieves substantially lower normalized densities and inverse densities, particularly at the 1\% (11.21 and 69.10). Moreover, Mini-Splatting consistently maintains lower normalized densities and inverse densities across the 1\%, 10\%, and 20\% ranges. This trend is observed consistently across all scans, suggesting that our approach effectively mitigates the clustering behavior of Gaussians while improving their image-space coverage. These findings corroborate the effectiveness of our method in enhancing the spatial distribution of Gaussians across scenes.

\subsection{Discussion on Distribution Desirability}\label{Sec:Discussion on Distribution Desirability}

A uniform distribution is desirable in regions with pronounced image gradient magnitudes to mitigate the irregular distribution introduced by 3DGS, while it is not necessarily desirable for regions with negligible image gradients.

\noindent\textbf{Distribution in Textured Regions.} In textured regions, characterized by pronounced image gradient magnitudes, 3DGS employs gradient-triggered densification, which often leads to clustered and redundant Gaussians as well as uncovered and blurry areas. 
In contrast, Mini-Splatting explicitly regulates the spatial distribution of Gaussians, achieving a more balanced and uniform distribution that is desirable for scene modeling.
By promoting a more balanced distribution, Mini-Splatting reduces local density in overlapping regions, eliminating clustered Gaussians and avoiding redundant computation, and simultaneously increases density in uncovered areas, improving scene coverage and mitigating blurry artifacts.

As illustrated in Figs. \ref{fig:density} and \ref{fig:inv_density}, our approach regulates the spatial organization of Gaussians, achieving a more balanced distribution and mitigating spatial irregularity. Figure \ref{fig:qual} demonstrates the improved image clarity, where blurry artifacts are significantly reduced. In addition, Table \ref{tab:quant} reports a 6.8$\times$ reduction in Gaussian count achieved by Mini-Splatting. These results indicate that enhancing distribution quality improves compactness while maintaining rendering fidelity.

\noindent\textbf{Distribution in Textureless Regions.} In textureless regions, characterized by negligible image gradients, a uniform distribution is not necessarily desirable. In such areas, a small number of large and flattened Gaussians are sufficient to capture both geometry and appearance. 
Enforcing uniformity in these regions would introduce redundant Gaussians \revise{and} increase memory consumption with limited performance \revise{gains}.

For instance, Figure \ref{fig:fail} visualizes the projected Gaussian centers of \textit{counter} and \textit{playroom} along with PSNR, normalized density, and normalized inverse density at top 10\% Gaussians and area. It is evident that Mini-Splatting achieves a more balanced distribution than 3DGS, exhibiting lower normalized density and inverse density. Despite its balanced distribution, Mini-Splatting yields lower PSNR values for certain test views (29.70~dB for Mini-Splatting vs. 30.42~dB for 3DGS on \textit{counter}), suggesting that fewer Gaussians are sufficient for textureless regions. These results indicate that improved scene coverage through a more uniform distribution may introduce unnecessary Gaussians and is therefore not beneficial for representation compactness in such cases.

}

\section{Conclusion}


In this work, we present Mini-Splatting2, a unified framework for efficient scene modeling with 3D Gaussians. To address irregularity in \revise{spatial distribution}, we propose a structure-aware distribution scheme comprising adaptive Gaussian organization and redundant Gaussian simplification, which explicitly regulates the spatial placement of Gaussians. To alleviate indiscrimination in training dynamics, we introduce a region-prioritized optimization scheme that incorporates aggressive model growth and occluded Gaussian culling, thereby guiding the optimization process toward geometrically informative regions of the scene. Extensive experiments demonstrate that Mini-Splatting2 achieves a favorable balance among representation compactness, optimization speed, and modeling quality. These findings establish a solid foundation for advancing efficient and high-quality 3D scene modeling within the paradigm of 3D Gaussian Splatting.

\bibliographystyle{IEEEtran}
\bibliography{ref}

@String(CVPR= {IEEE Conf. Comput. Vis. Pattern Recog.})

@String(ICCV= {Int. Conf. Comput. Vis.})

@String(ECCV= {Eur. Conf. Comput. Vis.})

@String(BMVC= {Brit. Mach. Vis. Conf.})

@String(TOG= {ACM Trans. Graph.})

@String(TVCG  = {IEEE Trans. Vis. Comput. Graph.})

@String(ICLR = {Int. Conf. Learn. Represent.})

@String(CVPRW= {IEEE Conf. Comput. Vis. Pattern Recog. Worksh.})

@String(CGF  = {Comput. Graph. Forum})

@String(CVPR  = {CVPR})

@String(ICCV  = {ICCV})

@String(ECCV  = {ECCV})

@String(BMVC  =	{BMVC})

@String(TOG   = {ACM TOG})

@String(TVCG  = {IEEE TVCG})

@String(ICLR  = {ICLR})

@String(CVPRW= {CVPRW})

@inproceedings{barron2021mipnerf,
      title={{Mip-NeRF: A Multiscale Representation for Anti-Aliasing Neural Radiance Fields}},
      author={Jonathan T. Barron and Ben Mildenhall and Matthew Tancik and Peter Hedman and Ricardo Martin-Brualla and Pratul P. Srinivasan},
      year={2021},
      booktitle={ICCV},
}

@inproceedings{barron2023zip,
  title={{Zip-NeRF: Anti-Aliased Grid-Based Neural Radiance Fields}},
  author={\vspace{0mm}Barron, Jonathan T and Mildenhall, Ben and Verbin, Dor and Srinivasan, Pratul P and Hedman, Peter},
      year={2023},
      booktitle={ICCV},
}

@inproceedings{mildenhall2020,
 title={{NeRF: Representing Scenes as Neural Radiance Fields for View Synthesis}},
 author={Ben Mildenhall and Pratul P. Srinivasan and Matthew Tancik and Jonathan T. Barron and Ravi Ramamoorthi and Ren Ng},
 year={2020},
 booktitle={ECCV},
}

@inproceedings{schoenberger2016sfm,
    author={Sch\"{o}nberger, Johannes Lutz and Frahm, Jan-Michael},
    title={{Structure-from-Motion Revisited}},
    booktitle={CVPR},
    year={2016},
}

@article{muller2022instant,
    title = {{Instant Neural Graphics Primitives with a Multiresolution Hash Encoding}},
  author={M{\"u}ller, Thomas and Evans, Alex and Schied, Christoph and Keller, Alexander},
  journal={ACM TOG},
  volume={41},
  number={4},
  pages={1--15},
  year={2022},
}

@article{hedman2018deep,
  title={{Deep Blending for Free-Viewpoint Image-Based Rendering}},
  author={Hedman, Peter and Philip, Julien and Price, True and Frahm, Jan-Michael and Drettakis, George and Brostow, Gabriel},
  journal={ACM TOG},
  volume={37},
  number={6},
  pages={1--15},
  year={2018},
}

@inproceedings{kopanas2021point,
  title={{Point-Based Neural Rendering with Per-View Optimization}},
  author={Kopanas, Georgios and Philip, Julien and Leimk{\"u}hler, Thomas and Drettakis, George},
  booktitle={CGF},
  year={2021},
}

@article{Knapitsch2017,
    author    = {Arno Knapitsch and Jaesik Park and Qian-Yi Zhou and Vladlen Koltun},
    title     = {{Tanks and Temples: Benchmarking Large-Scale Scene Reconstruction}},
    journal   = {ACM TOG},
    volume    = {36},
    number    = {4},
    year      = {2017},
}

@inproceedings{barron2022mipnerf360,
    title={{Mip-{NeRF} 360: Unbounded Anti-Aliased Neural Radiance Fields}},
    author={Jonathan T. Barron and Ben Mildenhall and 
            Dor Verbin and Pratul P. Srinivasan and Peter Hedman},
    booktitle={CVPR},
    year={2022}
}

@inproceedings{yu2021plenoxels,
  title={{Plenoxels: Radiance Fields without Neural Networks}},
  author={Yu, Alex and Fridovich-Keil, Sara and Tancik, Matthew and Chen, Qinhong and Recht, Benjamin and Kanazawa, Angjoo},
  booktitle={CVPR},
  year={2022}
}

@inproceedings{sun2022direct,
  title={{Direct Voxel Grid Optimization: Super-Fast Convergence for Radiance Fields Reconstruction}},
  author={Sun, Cheng and Sun, Min and Chen, Hwann-Tzong},
  booktitle={CVPR},
  year={2022}
}

@article{kerbl20233d,
  title={{3D Gaussian Splatting for Real-Time Radiance Field Rendering}},
  author={Kerbl, Bernhard and Kopanas, Georgios and Leimk{\"u}hler, Thomas and Drettakis, George},
  journal={ACM TOG},
  volume={42},
  number={4},
  pages={139--1},
  year={2023}
}

@inproceedings{niedermayr2024compressed,
  title={{Compressed 3D Gaussian Splatting for Accelerated Novel View Synthesis}},
  author={Niedermayr, Simon and Stumpfegger, Josef and Westermann, R{\"u}diger},
  booktitle={CVPR},
  year={2024}
}

@inproceedings{lee2024compact,
  title={{Compact 3D Gaussian Representation for Radiance Field}},
  author={Lee, Joo Chan and Rho, Daniel and Sun, Xiangyu and Ko, Jong Hwan and Park, Eunbyung},
  booktitle={CVPR},
  year={2024}
}

@inproceedings{li2023compressing,
  title={{Compressing Volumetric Radiance Fields to 1 MB}},
  author={Li, Lingzhi and Shen, Zhen and Wang, Zhongshu and Shen, Li and Bo, Liefeng},
  booktitle={CVPR},
  year={2023}
}

@inproceedings{fan2024lightgaussian,
  title={{LightGaussian: Unbounded 3D Gaussian Compression with 15x Reduction and 200+ FPS}},
  author={Fan, Zhiwen and Wang, Kevin and Wen, Kairun and Zhu, Zehao and Xu, Dejia and Wang, Zhangyang},
  booktitle={NeurIPS},
  year={2024}
}

@article{zwicker2002ewa,
  title={{EWA Splatting}},
  author={Zwicker, Matthias and Pfister, Hanspeter and Van Baar, Jeroen and Gross, Markus},
  journal={TVCG},
  volume={8},
  number={3},
  pages={223--238},
  year={2002},
  publisher={IEEE}
}

@inproceedings{xu2022point,
  title={{Point-NeRF: Point-Based Neural Radiance Fields}},
  author={Xu, Qiangeng and Xu, Zexiang and Philip, Julien and Bi, Sai and Shu, Zhixin and Sunkavalli, Kalyan and Neumann, Ulrich},
  booktitle={CVPR},
  year={2022}
}

@inproceedings{deng2022depth,
  title={{Depth-Supervised NeRF: Fewer Views and Faster Training for Free}},
  author={Deng, Kangle and Liu, Andrew and Zhu, Jun-Yan and Ramanan, Deva},
  booktitle={CVPR},
  year={2022}
}

@inproceedings{tang2023dreamgaussian,
  title={{DreamGaussian: Generative Gaussian Splatting for Efficient 3D Content Creation}},
  author={Tang, Jiaxiang and Ren, Jiawei and Zhou, Hang and Liu, Ziwei and Zeng, Gang},
  booktitle={ICLR},
  year={2024}
}

@inproceedings{franke2024trips,
  title={{TRIPS: Trilinear Point Splatting for Real-Time Radiance Field Rendering}},
  author={Franke, Linus and R{\"u}ckert, Darius and Fink, Laura and Stamminger, Marc},
  booktitle={CGF},
  year={2024},
}

@inproceedings{yu2023mip,
  title={{Mip-Splatting: Alias-Free 3D Gaussian Splatting}},
  author={Yu, Zehao and Chen, Anpei and Huang, Binbin and Sattler, Torsten and Geiger, Andreas},
  booktitle={CVPR},
  year={2024}
}

@inproceedings{deng2023compressing,
  title={{Compressing Explicit Voxel Grid Representations: Fast NeRFs Become Also Small}},
  author={Deng, Chenxi Lola and Tartaglione, Enzo},
  booktitle={WACV},
  year={2023}
}

@inproceedings{xie2023hollownerf,
  title={{HollowNeRF: Pruning Hashgrid-Based NeRFs with Trainable Collision Mitigation}},
  author={Xie, Xiufeng and Gherardi, Riccardo and Pan, Zhihong and Huang, Stephen},
  booktitle={ICCV},
  year={2023}
}

@inproceedings{rho2023masked,
  title={{Masked Wavelet Representation for Compact Neural Radiance Fields}},
  author={Rho, Daniel and Lee, Byeonghyeon and Nam, Seungtae and Lee, Joo Chan and Ko, Jong Hwan and Park, Eunbyung},
  booktitle={CVPR},
  year={2023}
}

@inproceedings{schoenberger2016mvs,
    author={Sch\"{o}nberger, Johannes Lutz and Zheng, Enliang and Pollefeys, Marc and Frahm, Jan-Michael},
    title={{Pixelwise View Selection for Unstructured Multi-View Stereo}},
    booktitle={ECCV},
    year={2016},
}

@inproceedings{taming3dgs,
  title={{Taming 3DGS: High-Quality Radiance Fields with Limited Resources}},
  author={Mallick, Saswat Subhajyoti and Goel, Rahul and Kerbl, Bernhard and Steinberger, Markus and Carrasco, Francisco Vicente and De La Torre, Fernando},
  booktitle={SIGGRAPH Asia},
  year={2024}
}

@inproceedings{fang2024mini,
    author={Fang, Guangchi and Wang, Bing},
    title={{Mini-Splatting: Representing Scenes with a Constrained Number of Gaussians}},
    booktitle={ECCV},
    year={2024},
}

@inproceedings{liu2020neural,
  title={{Neural Sparse Voxel Fields}},
  author={Liu, Lingjie and Gu, Jiatao and Zaw Lin, Kyaw and Chua, Tat-Seng and Theobalt, Christian},
  booktitle={NeurIPS},
  year={2020}
}

@inproceedings{Huang2DGS2024,
    title={{2D Gaussian Splatting for Geometrically Accurate Radiance Fields}},
    author={Huang, Binbin and Yu, Zehao and Chen, Anpei and Geiger, Andreas and Gao, Shenghua},
    booktitle = {SIGGRAPH},
    year      = {2024},
}

@article{Yu2024GOF,
  title     = {{Gaussian Opacity Fields: Efficient Adaptive Surface Reconstruction in Unbounded Scenes}},
  author={Yu, Zehao and Sattler, Torsten and Geiger, Andreas},
  journal={ACM TOG},
  volume={43},
  number={6},
  pages={1--13},
  year={2024},
}

@inproceedings{LaRa,
         author = {Anpei Chen and Haofei Xu and Stefano Esposito and Siyu Tang and Andreas Geiger},
         title = {{LaRa: Efficient Large-Baseline Radiance Fields}},
         booktitle = {ECCV},
         year = {2024},
        }

@inproceedings{matsuki2024gaussian,
  title={{Gaussian Splatting Slam}},
  author={Matsuki, Hidenobu and Murai, Riku and Kelly, Paul HJ and Davison, Andrew J},
  booktitle={CVPR},
  year={2024}
}

@inproceedings{yan2024gs,
  title={{GS-SLAM: Dense Visual SLAM with 3D Gaussian Splatting}},
  author={Yan, Chi and Qu, Delin and Xu, Dan and Zhao, Bin and Wang, Zhigang and Wang, Dong and Li, Xuelong},
  booktitle={CVPR},
  year={2024}
}

@inproceedings{zhang2024pixelgs,
  title     = {{Pixel-GS: Density Control with Pixel-Aware Gradient for 3D Gaussian Splatting}},
  author    = {Zhang, Zheng and Hu, Wenbo and Lao, Yixing and He, Tong and Zhao, Hengshuang},
  booktitle = {ECCV},
  year      = {2024}
}

@inproceedings{ye2024absgs,
  title={{AbsGS: Recovering Fine Details for 3D Gaussian Splatting}},
  author={Ye, Zongxin and Li, Wenyu and Liu, Sidun and Qiao, Peng and Dou, Yong},
  booktitle={ACM MM},
  year={2024}
}

@inproceedings{bulo2024revising,
  title={{Revising Densification in Gaussian Splatting}},
  author={Bul{\`o}, Samuel Rota and Porzi, Lorenzo and Kontschieder, Peter},
  booktitle = {ECCV},
  year={2024}
}

@inproceedings{kim2024color,
  title={{Color-Cued Efficient Densification Method for 3D Gaussian Splatting}},
  author={Kim, Sieun and Lee, Kyungjin and Lee, Youngki},
  booktitle={CVPRW},
  year={2024}
}

@inproceedings{cheng2024gaussianpro,
  title={{GaussianPro: 3D Gaussian Splatting with Progressive Propagation}},
  author={Cheng, Kai and Long, Xiaoxiao and Yang, Kaizhi and Yao, Yao and Yin, Wei and Ma, Yuexin and Wang, Wenping and Chen, Xuejin},
  booktitle={ICML},
  year={2024}
}

@article{li2025mvg,
  title={{MVG-Splatting: Multi-View Guided Gaussian Splatting with Adaptive Quantile-Based Geometric Consistency Densification}},
  author={Li, Zhuoxiao and Yao, Shanliang and Chu, Yijie and Garcia-Fernandez, Angel F and Yue, Yong and Ding, Weiping and Zhu, Xiaohui},
  journal={Information Fusion},
  volume = {126},
  pages = {103540},
  year={2025},
}

@article{du2024mvgs,
  title={{MVGS: Multi-View-Regulated Gaussian Splatting for Novel View Synthesis}},
  author={Du, Xiaobiao and Wang, Yida and Yu, Xin},
  journal={arXiv preprint arXiv:2410.02103},
  year={2024}
}

@inproceedings{kheradmand20243d,
    title={{3D Gaussian Splatting as Markov Chain Monte Carlo}},
  author={Kheradmand, Shakiba and Rebain, Daniel and Sharma, Gopal and Sun, Weiwei and Tseng, Jeff and Isack, Hossam and Kar, Abhishek and Tagliasacchi, Andrea and Yi, Kwang Moo},
    booktitle={NeurIPS},
    year={2024}
}

@article{fan2024instantsplat,
  title={{InstantSplat: Unbounded Sparse-View Pose-Free Gaussian Splatting in 40 Seconds}}, 
  author={Fan, Zhiwen and Cong, Wenyan and Wen, Kairun and Wang, Kevin and Zhang, Jian and Ding, Xinghao and Xu, Danfei and Ivanovic, Boris and Pavone, Marco and Pavlakos, Georgios and Wang, Zhangyang and Wang, Yue},
  journal={arXiv preprint arXiv:2403.20309},
  year={2024}
}

@inproceedings{charatan23pixelsplat,
      title={{PixelSplat: 3D Gaussian Splats from Image Pairs for Scalable Generalizable 3D Reconstruction}},
      author={David Charatan and Sizhe Li and Andrea Tagliasacchi and Vincent Sitzmann},
      year={2024},
      booktitle={CVPR},
}

@inproceedings{chen2024mvsplat,
    title   = {{MVSplat: Efficient 3D Gaussian Splatting from Sparse Multi-View Images}},
    author  = {Chen, Yuedong and Xu, Haofei and Zheng, Chuanxia and Zhuang, Bohan and Pollefeys, Marc and Geiger, Andreas and Cham, Tat-Jen and Cai, Jianfei},
    booktitle={ECCV},
    year    = {2024},
}

@inproceedings{liu2025mvsgaussian,
  title={{MVSGaussian: Fast Generalizable Gaussian Splatting Reconstruction from Multi-View Stereo}},
  author={Liu, Tianqi and Wang, Guangcong and Hu, Shoukang and Shen, Liao and Ye, Xinyi and Zang, Yuhang and Cao, Zhiguo and Li, Wei and Liu, Ziwei},
  booktitle={ECCV},
    year    = {2024},
}

@article{szymanowicz2024flash3d,
  title={{Flash3D: Feed-Forward Generalisable 3D Scene Reconstruction from a Single Image}},
  author={Szymanowicz, Stanislaw and Insafutdinov, Eldar and Zheng, Chuanxia and Campbell, Dylan and Henriques, Jo{\~a}o F and Rupprecht, Christian and Vedaldi, Andrea},
  journal={arXiv preprint arXiv:2406.04343},
  year={2024}
}

@article{durvasula2023distwar,
  title={{DISTWAR: Fast Differentiable Rendering on Raster-Based Rendering Pipelines}},
  author={Durvasula, Sankeerth and Zhao, Adrian and Chen, Fan and Liang, Ruofan and Sanjaya, Pawan Kumar and Vijaykumar, Nandita},
  journal={arXiv preprint arXiv:2401.05345},
  year={2023}
}

@article{ye2024gsplat,
  title={{Gsplat: An Open-Source Library for Gaussian Splatting}},
  author={Ye, Vickie and Li, Ruilong and Kerr, Justin and Turkulainen, Matias and Yi, Brent and Pan, Zhuoyang and Seiskari, Otto and Ye, Jianbo and Hu, Jeffrey and Tancik, Matthew and Kanazawa, Angjoo},
  journal={JMLR},
  volume={26},
  number={34},
  pages={1--17},
  year={2025}
}

@article{radl2024stopthepop,
  title={{StopThePop: Sorted Gaussian Splatting for View-Consistent Real-Time Rendering}},
  author={Radl, Lukas and Steiner, Michael and Parger, Mathias and Weinrauch, Alexander and Kerbl, Bernhard and Steinberger, Markus},
  journal={ACM TOG},
  volume={43},
  number={4},
  pages={1--17},
  year={2024},
}

@inproceedings{hollein20253dgs,
  title={{3DGS-LM: Faster Gaussian-Splatting Optimization with Levenberg-Marquardt}},
  author={H{\"o}llein, Lukas and Bo{\v{z}}i{\v{c}}, Alja{\v{z}} and Zollh{\"o}fer, Michael and Nie{\ss}ner, Matthias},
  booktitle={CVPR},
  year={2025}
}

@inproceedings{zhang2018unreasonable,
  title={{The Unreasonable Effectiveness of Deep Features as a Perceptual Metric}},
  author={Zhang, Richard and Isola, Phillip and Efros, Alexei A and Shechtman, Eli and Wang, Oliver},
  booktitle={CVPR},
  year={2018}
}

@article{mai2024ever,
  title={{EVER: Exact Volumetric Ellipsoid Rendering for Real-Time View Synthesis}},
  author={Mai, Alexander and Hedman, Peter and Kopanas, George and Verbin, Dor and Futschik, David and Xu, Qiangeng and Kuester, Falko and Barron, Jonathan T and Zhang, Yinda},
  journal={arXiv preprint arXiv:2410.01804},
  year={2024}
}

@inproceedings{navaneet2024compgs,
  title={{CompGS: Smaller and Faster Gaussian Splatting with Vector Quantization}},
  author={Navaneet, KL and Pourahmadi Meibodi, Kossar and Abbasi Koohpayegani, Soroush and Pirsiavash, Hamed},
  booktitle={ECCV},
  year={2024},
}

@inproceedings{ali2024trimming,
  title={{Trimming the Fat: Efficient Compression of 3D Gaussian Splats through Pruning}},
  author={Ali, Muhammad Salman and Qamar, Maryam and Bae, Sung-Ho and Tartaglione, Enzo},
  booktitle={BMVC},
  year={2024}
}

@inproceedings{lu2024scaffold,
  title={{Scaffold-GS: Structured 3D Gaussians for View-Adaptive Rendering}},
  author={Lu, Tao and Yu, Mulin and Xu, Linning and Xiangli, Yuanbo and Wang, Limin and Lin, Dahua and Dai, Bo},
  booktitle={CVPR},
  year={2024}
}

@inproceedings{chen2024hac,
  title={{HAC: Hash-Grid Assisted Context for 3D Gaussian Splatting Compression}},
  author={Chen, Yihang and Wu, Qianyi and Lin, Weiyao and Harandi, Mehrtash and Cai, Jianfei},
  booktitle={ECCV},
  year={2024},
}

@inproceedings{wang2024contextgs,
  title={{ContextGS: Compact 3D Gaussian Splatting with Anchor Level Context Model}},
  author={Wang, Yufei and Li, Zhihao and Guo, Lanqing and Yang, Wenhan and Kot, Alex and Wen, Bihan},
  booktitle={NeurIPS},
  year={2024}
}

@article{kerbl2024hierarchical,
  title={{A Hierarchical 3D Gaussian Representation for Real-Time Rendering of Very Large Datasets}},
  author={Kerbl, Bernhard and Meuleman, Andreas and Kopanas, Georgios and Wimmer, Michael and Lanvin, Alexandre and Drettakis, George},
  journal={ACM TOG},
  volume={43},
  number={4},
  pages={1--15},
  year={2024},
}

@article{ren2024octree,
  title={{Octree-GS: Towards Consistent Real-Time Rendering with LOD-Structured 3D Gaussians}},
  author={Ren, Kerui and Jiang, Lihan and Lu, Tao and Yu, Mulin and Xu, Linning and Ni, Zhangkai and Dai, Bo},
  journal={arXiv preprint arXiv:2403.17898},
  year={2024}
}

@inproceedings{liu2024citygaussian,
  title={{CityGaussian: Real-Time High-Quality Large-Scale Scene Rendering with Gaussians}},
  author={Liu, Yang and Luo, Chuanchen and Fan, Lue and Wang, Naiyan and Peng, Junran and Zhang, Zhaoxiang},
  booktitle={ECCV},
  year={2024},
}

@article{lu2024turbo,
  title={{Turbo-GS: Accelerating 3D Gaussian Fitting for High-Quality Radiance Fields}},
  author={Lu, Tao and Dhiman, Ankit and Srinath, R and Arslan, Emre and Xing, Angela and Xiangli, Yuanbo and Babu, R Venkatesh and Sridhar, Srinath},
  journal={arXiv preprint arXiv:2412.13547},
  year={2024}
}

@inproceedings{zhang2025gaussianspa,
  title={{GaussianSpa: An Optimizing-Sparsifying Simplification Framework for Compact and High-Quality 3D Gaussian Splatting}},
  author={Zhang, Yangming and Jia, Wenqi and Niu, Wei and Yin, Miao},
  booktitle={CVPR},
  year={2025}
}

@inproceedings{chen2025dashgaussian,
  title={{DashGaussian: Optimizing 3D Gaussian Splatting in 200 Seconds}},
  author={Chen, Youyu and Jiang, Junjun and Jiang, Kui and Tang, Xiao and Li, Zhihao and Liu, Xianming and Nie, Yinyu},
  booktitle={CVPR},
  year={2025}
}

@article{song2025adgaussian,
  title={{ADGaussian: Generalizable Gaussian Splatting for Autonomous Driving with Multi-modal Inputs}},
  author={Song, Qi and Li, Chenghong and Lin, Haotong and Peng, Sida and Huang, Rui},
  journal={arXiv preprint arXiv:2504.00437},
  year={2025}
}

@inproceedings{zhou2024drivinggaussian,
  title={{Drivinggaussian: Composite Gaussian Splatting for Surrounding Dynamic Autonomous Driving Scenes}},
  author={Zhou, Xiaoyu and Lin, Zhiwei and Shan, Xiaojun and Wang, Yongtao and Sun, Deqing and Yang, Ming-Hsuan},
  booktitle={CVPR},
  year={2024}
}

\end{document}